\documentclass{ieeeaccess}

\usepackage{cite}
\usepackage{amsmath,amssymb,amsfonts}
\usepackage{algorithmic}
\usepackage{graphicx}
\usepackage{textcomp}
\def\BibTeX{{\rm B\kern-.05em{\sc i\kern-.025em b}\kern-.08em
        T\kern-.1667em\lower.7ex\hbox{E}\kern-.125emX}}
    
\usepackage[figuresright]{rotating} % Rotating tables & figures
\usepackage{booktabs}
\usepackage{array}
\usepackage{multirow}
\usepackage{xcolor}
\newcommand{\etal}{\textit{et al.}}

\begin{document}
    
\history{Date of publication xxxx 00, 0000, date of current version xxxx 00, 0000.}
\doi{10.1109/ACCESS.2017.DOI}

\title{A Weakly-Supervised Semantic Segmentation Approach based on the Centroid Loss: Application to Quality Control and Inspection}
\author{\uppercase{Kai Yao}\authorrefmark{1}, 
        \uppercase{Alberto Ortiz}\authorrefmark{1}, \IEEEmembership{Member, IEEE}, and 
        \uppercase{Francisco Bonnin-Pascual}\authorrefmark{1}}
\address[1]{Department of Mathematics and Computer Science (University of the Balearic Islands) and IDISBA (Institut d'Investigacio Sanitaria de les Illes Balears), Palma de Mallorca, Spain; \{kai.yao, alberto.ortiz, xisco.bonnin\}@uib.es}
\tfootnote{This work is partially supported by EU-H2020 projects BUGWRIGHT2 (GA 871260) and ROBINS (GA 779776), PGC2018-095709-B-C21 (MCIU/AEI/FEDER, UE), and PROCOE/4/2017 (Govern Balear, 50\% P.O. FEDER 2014-2020 Illes Balears). This publication reflects only the authors views and the European Union is not liable for any use that may be made of the information contained therein.}

% Kai Yao \headeretal
\markboth
{K. Yao, A. Ortiz and F. Bonnin-Pascual: A Weakly-Supervised Semantic Segmentation Approach based on the Centroid Loss}
{K. Yao, A. Ortiz and F. Bonnin-Pascual: A Weakly-Supervised Semantic Segmentation Approach based on the Centroid Loss}

\corresp{Corresponding author: Alberto Ortiz (e-mail: alberto.ortiz@uib.es).}

\begin{abstract}
It is generally accepted that one of the critical parts of current vision algorithms based on deep learning and convolutional neural networks is the annotation of a sufficient number of images to achieve competitive performance. This is particularly difficult for semantic segmentation tasks since the annotation must be ideally generated at the pixel level. Weakly-supervised semantic segmentation aims at reducing this cost by employing simpler annotations that, hence, are easier, cheaper and quicker to produce. In this paper, we propose and assess a new weakly-supervised semantic segmentation approach making use of a novel loss function whose goal is to counteract the effects of weak annotations. To this end, this loss function comprises several terms based on partial cross-entropy losses, being one of them the Centroid Loss. This term induces a clustering of the image pixels in the object classes under consideration, whose aim is to improve the training of the segmentation network by guiding the optimization. The performance of the approach is evaluated against datasets from two different industry-related case studies: while one involves the detection of instances of a number of different object classes in the context of a quality control application, the other stems from the visual inspection domain and deals with the localization of images areas whose pixels correspond to scene surface points affected by a specific sort of defect. The detection results that are reported for both cases show that, despite the differences among them and the particular challenges, the use of weak annotations do not prevent from achieving a competitive performance level for both. 
\end{abstract}

\begin{keywords}
Object Recognition, Quality Control and Inspection, Weakly-Supervised Semantic Segmentation
\end{keywords} 

\titlepgskip=-15pt

\maketitle

\section{Introduction}
\label{sc:introduction}

\PARstart{I}{mage} segmentation is a classical problem in computer vision aiming at distinguishing meaningful units in processed images. To this end, image pixels are grouped into regions that on many occasions are expected to correspond to the scene object projections. One step further identifies each unit as belonging to a particular class among a set of object classes to be recognized, giving rise to the Multi-Class Semantic Segmentation (MCSS) problem. From classical methods (e.g. region growing~\cite{Gonzalez2018}) to more robust methods (e.g. level-set~\cite{Wang2020} and graph-cut~\cite{Boykov2006}), various techniques have been proposed to achieve automatic image segmentation in a wide range of problems. Nevertheless, it has not been until recently that the performance of image segmentation algorithms has attained truly competitive levels, and this has been mostly thanks to the power of machine learning-based methodologies.

On the basis of the concept of Convolutional Neural Networks (CNN) proposed by LeCun and his collaborators (e.g. in the form of the well-known LeNet networks~\cite{LeCun1998}) and followed by the technological breaktrough that allowed training artificial neural structures with a number of parameters amounting to millions~\cite{Krizhevsky2012}, deep CNNs have demonstrated remarkable capabilities for problems so complex as image classification, multi-instance multi-object detection or multi-class semantic segmentation. And all this has been accomplished because of the "learning the representation" capacity of CNNs, embedded in the set of multi-scale feature maps defined in their architecture through non-linear activation functions and a number of convolutional filters that are automatically learnt during the training process by means of iterative back-propagation of prediction errors between current and expected output.

Regarding DCNN-based image segmentation, Guo et al.~\cite{Guo2018} distinguish among three categories of MCSS approaches in accordance to the methodology adopted while dealing with the input images (and correspondingly the required network structure): region-based semantic segmentation, semantic segmentation based on Fully Convolutional Networks (FCN) and Weakly-Supervised semantic segmentation (WSSS). While the former follows a \textit{segmentation using recognition} pipeline, which first detects free-form image regions, and next describes and classifies them, the second approach adopts a \textit{pixel-to-pixel map learning} strategy as key idea without resorting to the image region concept, and, lastly, WSSS methods focus on achieving a performance level similar to that of Fully-Supervised methods (FSSS) but with a weaker labelling of the training image set, i.e. less spatially-informative annotations than the pixel level, to simplify the generation of ground truth data. It is true that powerful interactive tools have been developed for annotating images at the pixel level, which, in particular, just require that the annotator draws a minimal polygon surrounding the targets (see e.g. the open annotation tool by the MIT~\cite{Labelme2016}). However, it still takes a few minutes on average to label the target areas for every picture (e.g. around 10 minutes on average for MS COCO labellers~\cite{Lin2014, Chan2020}), what makes WSSS methods interesting by themselves and actually quite convenient in general.

In this work, we focus on this last class of methods and propose a novel WSSS strategy based on a new loss function combining several terms to counteract the simplicity of the annotations. The strategy is, in turn, evaluated through a benchmark comprising two industry-related application cases of a totally different nature. One of these cases involves the detection of instances of a number of different object classes in the context of a \textit{quality control} application, while the other stems from the \textit{visual inspection} domain and deals with the detection of irregularly-shaped sets of pixels corresponding to scene defective areas. The details about the two application cases can be found in Section~\ref{sc:scenarios}. 

\begin{figure}
    \centering
    \begin{tabular}{@{\hspace{0mm}}c@{\hspace{0mm}}c@{\hspace{0mm}}c@{\hspace{0mm}}c@{\hspace{0mm}}}
        \includegraphics[width=0.25\columnwidth]{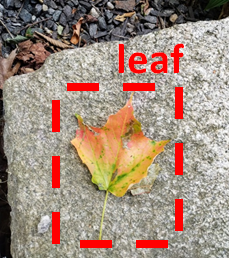} &
        \includegraphics[width=0.25\columnwidth]{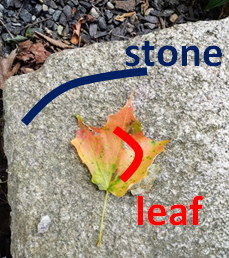} &
        \includegraphics[width=0.25\columnwidth]{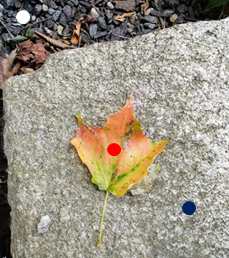} &
        \includegraphics[width=0.25\columnwidth]{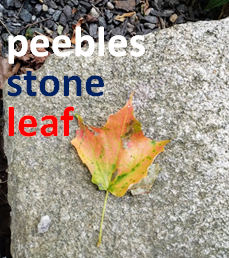} \\
        \footnotesize (a) & \footnotesize (b) & \footnotesize (c) & \footnotesize (d)
    \end{tabular}
    \caption{Examples of weak annotations, from more to less informative: (a) bounding boxes, (b) scribbles, (c) point-level labels, (d) image-level labels.}
    \label{fg:weak-annotations}
\end{figure}

WSSS methods are characterized, among others, by the sort of weak annotation that is assumed. In this regard, Chan \etal{} highlight in~\cite{Chan2020} several weak annotation methodologies, namely bounding boxes, scribbles, image points and image-level labels (see Fig.~\ref{fg:weak-annotations} for an illustration of all of them). In this work, we adopt a scribble-based methodology from which training masks are derived to propagate the category information from the labelled pixels to the unlabelled pixels during network training. 

The main contributions of this work are summarized as follows: %\re{ELABORATE MORE}
\begin{itemize}
    \item A new loss function $L$ comprising several partial cross entropy terms is developed to account for the vagueness of the annotations and the inherent noise of the training masks that are derived from them. This function includes a cluster centroids-based loss term, named as the Centroid Loss, which integrates a clustering process within the semantic segmentation approach.
    \item Another term of $L$ is defined through a Mean Squared Error (MSE) loss that cooperates with the other partial cross-entropy losses to refine the segmentation results.
    \item The Centroid Loss is embedded over a particular implementation of Attention U-Net~\cite{oktay2018attention}.
    \item We assess the performance of the full approach on a benchmark comprising two industry-related applications connected with, respectively, quality control and visual inspection.
\end{itemize}

A preliminary version of this work can be found in \cite{yao2020centroid} as a work-in progress paper.

The rest of the paper is organized as follows: Section~\ref{sc:scenarios} describes the two application scenarios we use as a benchmark of the semantic segmentation approach; Section~\ref{sc:related_work} reviews previous works on WSSS; Section~\ref{sc:methodology} describes the weakly-supervised methodology developed in this work; Section~\ref{sc:experiments} reports on the results of a number of experiments aiming at showing the performance of our approach from different points of view; and Section~\ref{sc:conclusions} concludes the paper and outlines future work.

% as well as deep learning approaches for inspection and quality control

\section{Application Scenarios}
\label{sc:scenarios}

\begin{figure}[t]  
    \centering
    \begin{tabular}{@{\hspace{0mm}}c@{\hspace{1mm}}c@{\hspace{1mm}}c@{\hspace{0mm}}}
        \includegraphics[width=0.31\columnwidth,height=0.31\columnwidth]{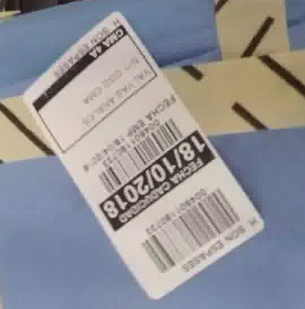}
        &
        \includegraphics[width=0.31\columnwidth,height=0.31\columnwidth]{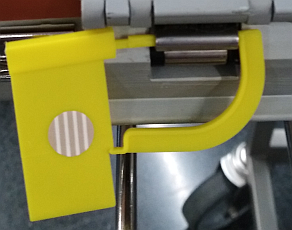}
        &
        \includegraphics[width=0.31\columnwidth,height=0.31\columnwidth]{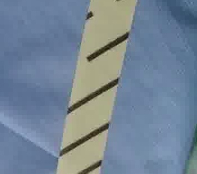} \\
        
        \includegraphics[width=0.31\columnwidth,height=0.31\columnwidth]{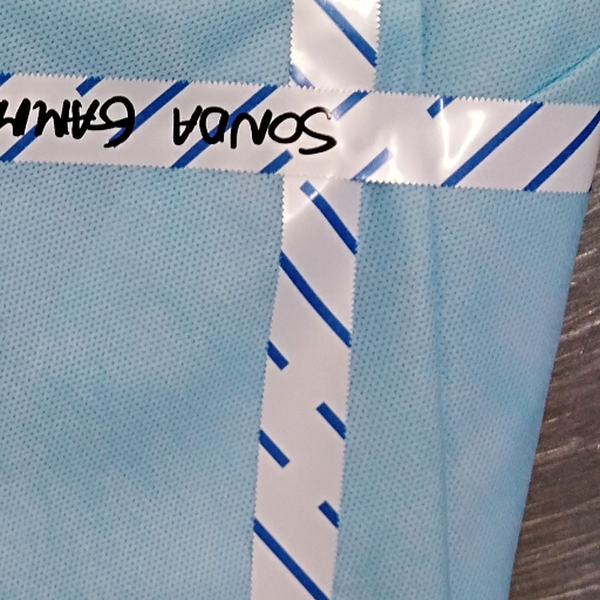}
        &
        \includegraphics[width=0.31\columnwidth,height=0.31\columnwidth]{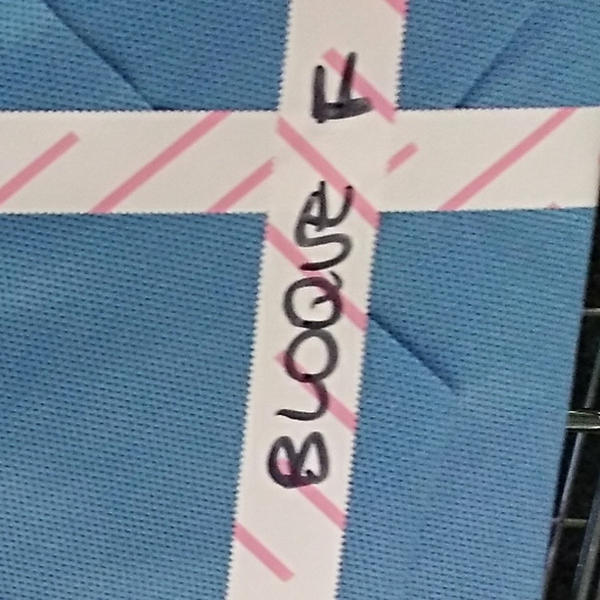}
        &
        \includegraphics[width=0.31\columnwidth,height=0.31\columnwidth]{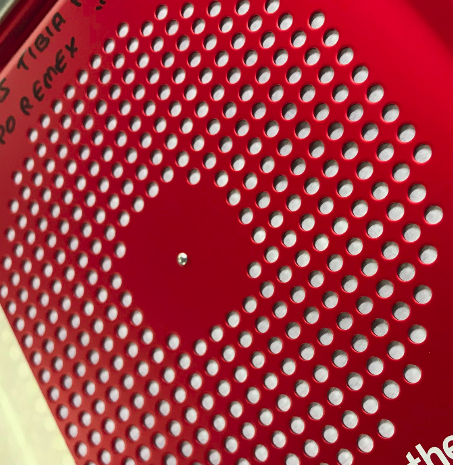} 
    \end{tabular}
    %\vspace{-2mm}
    \caption{Targets to be detected in the quality-control application considered in this work: from left to right and from top to bottom, box/bag tracking label, yellowish seal, three kinds of paper tape (black-, blue-, and pink-stripped) and white-dotted texture related to the presence of a whitish internal filter. All these elements, except the label, aim at evidencing the sterilization of the surgery tools contained in the box.}
    \label{fig:targets_qc}
\end{figure}

\begin{figure}[t]  
    \centering
    \begin{tabular}{@{\hspace{0mm}}c@{\hspace{1mm}}c@{\hspace{1mm}}c@{\hspace{0mm}}}
        \includegraphics[width=0.31\columnwidth,height=0.31\columnwidth]{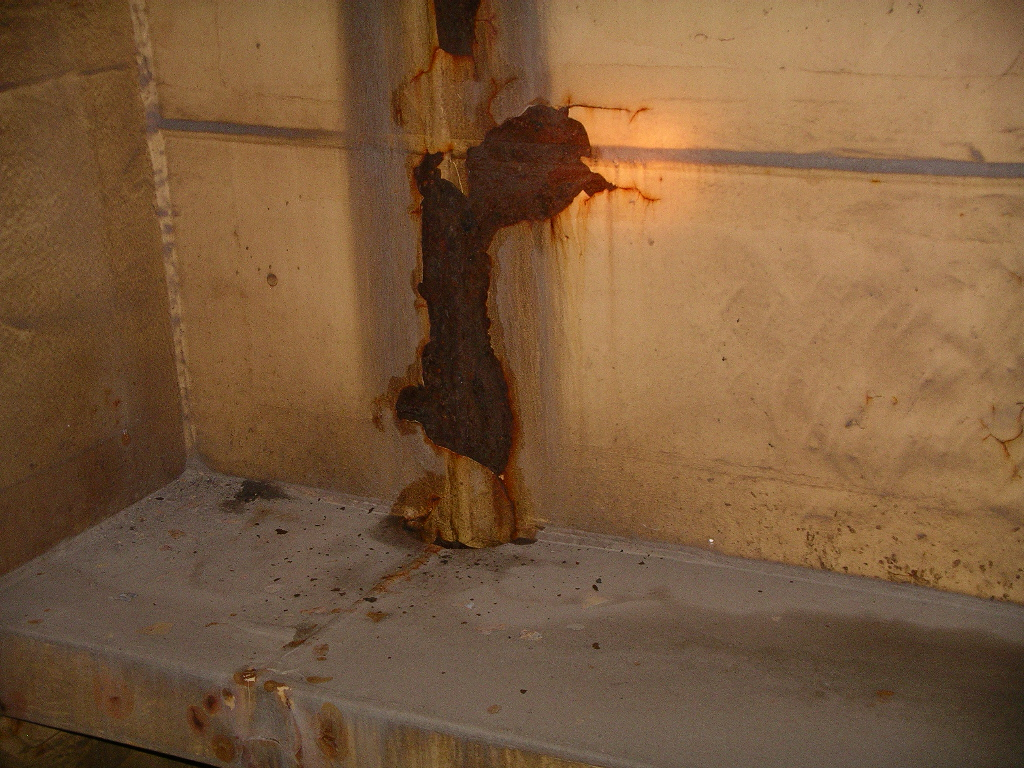}
        &
        \includegraphics[width=0.31\columnwidth,height=0.31\columnwidth]{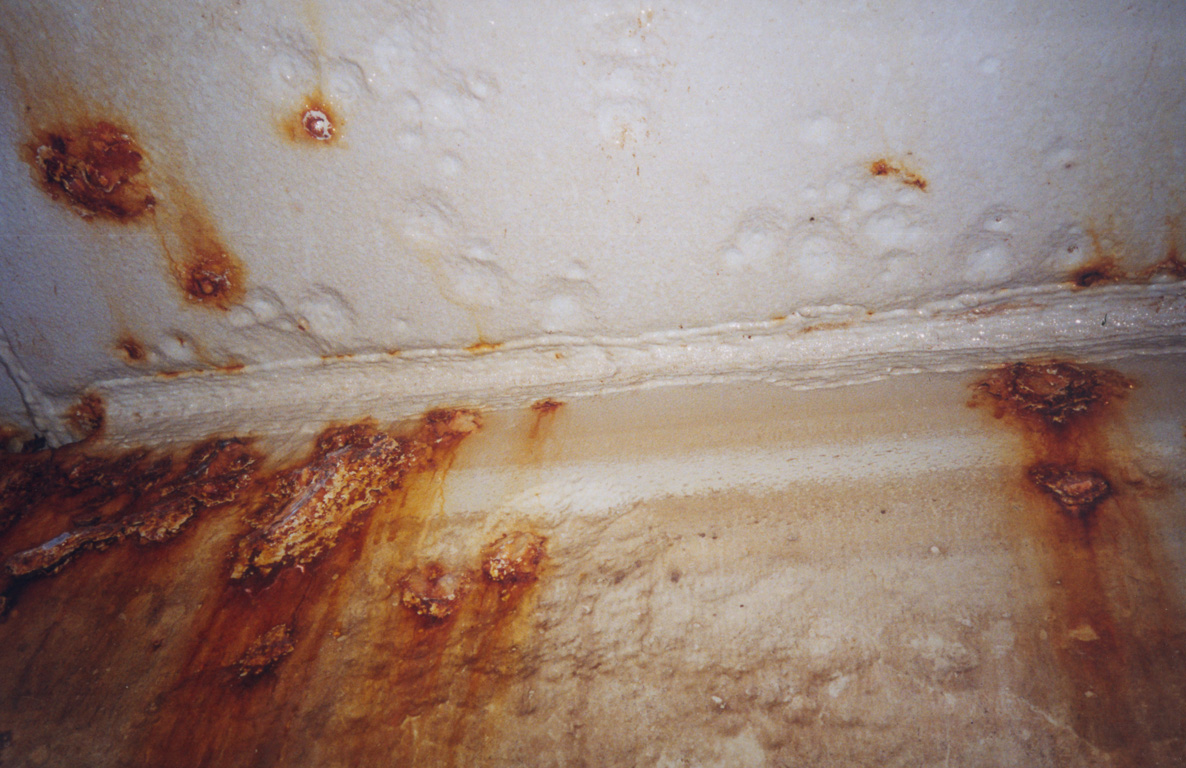}
        &
        \includegraphics[width=0.31\columnwidth,height=0.31\columnwidth]{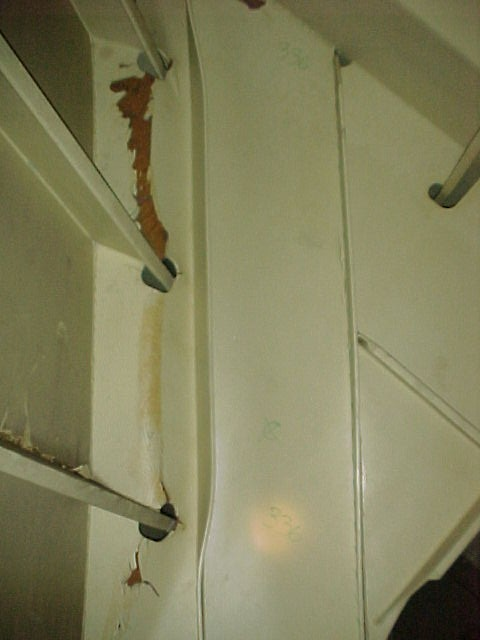}
    \end{tabular}
    %\vspace{-2mm}
    \caption{Targets to be detected in the visual inspection application considered in this work. The different images show examples of coating breakdown and corrosion affecting ship surfaces.}
    \label{fig:targets_insp}
\end{figure}

In this work, we use the two following industry-related application cases as a benchmark of the WSSS strategy that is developed: 
\begin{itemize}
    \item In the first case, we deal with the detection of a number of control elements that the sterilization unit of a hospital places in boxes and bags containing surgical tools that surgeons and nurses have to be supplied with prior to starting surgery. These elements provide evidence that the tools have been properly submitted to the required cleaning processes, i.e. they have been placed long enough inside the autoclave at a certain temperature, what makes them change their appearance. Figure~\ref{fig:targets_qc} shows, from left to right and top to bottom, examples of six kinds of elements to be detected for this application: the label/bar code used to track a box/bag of tools, the yellowish seal, the three kinds of paper tape which show the black-, blue- and pink-stripped appearance that can be observed in the figure, and an internal filter which is placed inside certain boxes and creates the white-dotted texture that can be noticed (instead of black-dotted when the filter is missing). All these elements, except the label, which is only for box/bag recording and tracking purposes, aim at corroborating the sterilization of the surgery tools contained in the box/bag. Finally, all of them may appear anywhere in the boxes/bags and in a different number, depending on the kind of box/bag.
    \item In the second case, we deal with the detection of one of the most common defects that can affect steel surfaces, i.e. coating breakdown and/or corrosion (CBC) in any of its many different forms. This is of particular relevance where the integrity of steel-based structures is critical, such as e.g. in large-tonnage vessels. An early detection, through suitable maintenance programmes, prevents vessel structures from suffering major damage which can ultimately compromise their integrity and lead to accidents with maybe catastrophic consequences for the crew (and passengers), environmental pollution or damage and/or total loss of the ship, its equipment and its cargo. The inspection of those ship-board structures by humans is a time-consuming, expensive and commonly hazardous activity, what, altogether, suggests the introduction of defect detection tools to alleviate the total cost of an inspection. Figure~\ref{fig:targets_insp} shows images of metallic vessel surfaces affected by CBC.
\end{itemize}
The quality control problem involves the detection of man-made, regular objects in a large variety of situations what leads to an important number of images to cover all cases, while the inspection problem requires the localization of image areas of irregular shape, and this makes harder and longer the labelling (specially for inexperienced staff). In both cases, it turns out to be relevant the use of a training methodology that reduces the cost of image annotation. As will be shown, our approach succeeds in both cases, despite the particular challenges and the differences among them, and the use of weak annotations do not prevent from achieving a competitive performance level for those particular problems (see also~\cite{Chan2020} for a discussion on some of the challenges the WSSS methods typically can have to face).

\section{Related Work}
\label{sc:related_work}

Although fully-supervised segmentation approaches based on DCNNs can achieve an excellent performance, they require plenty of pixel-wise annotations, what turns out to be very costly in practice. In response to this fact, researchers have recently paid attention to the use of weak annotations for addressing MCSS problems. Among others, \cite{wei2017object,choe2019attention,kolesnikov2016seed,yang2020combinational} consider image-level labels, while~\cite{khoreva2017simple,papandreou2015weakly} assume the availability of bounding boxes-based annotations, and~\cite{boykov2001interactive,grady2006random,bai2009geodesic,tang2018regularized,kervadec2019constrained,lin2016scribblesup,tang2018normalized,xu2015learning} make use of scribbles.

In more detail, most WSSS methods involving the use of image level-labels are based on the so-called Class Activation Maps (CAMs)~\cite{wei2017object}, as obtained from a classification network. In~\cite{choe2019attention}, an Attention-based Dropout Layer (ADL) is developed to obtain the entire outline of the target from the CAMs. The ADL relies on a self-attention mechanism to process the feature maps. More precisely, this layer is intended for a double purpose: one task is to hide the most discriminating parts in the feature maps, what induces the model to learn also the less discriminative part of the target, while the other task intends to highlight the informative region of the target for improving the recognition ability of the model. A three-term loss function is proposed in~\cite{kolesnikov2016seed} to seed, expand, and constrain the object regions progressively when the network is trained. Their loss function is based on three guiding principles: seed with weak localization cues, expand objects based on the information about which classes can occur in an image, and constrain the segmentation so as to coincide with object boundaries. In~\cite{yang2020combinational}, the focus is placed on how to obtain better CAMs. To this end, the work aims at solving the incorrect high response in CAMs through a linear combination of higher-dimensional CAMs.  

Regarding the use of bounding boxes as weakly-supervised annotations for semantic segmentation, in~\cite{khoreva2017simple}, GraphCut and Holistically-nested Edge Detection (HED) algorithms are combined to refine the bounding boxes ground truth and make predictions, while the refined ground truth is used to train the network iteratively. Similarly, in~\cite{papandreou2015weakly}, the authors develop a WSSS model using bounding boxes-based annotations, where: firstly, a segmentation network based on the DeepLab-CRF model obtains a series of coarse segmentation results, and, secondly, a dense Conditional Random Field (CRF)-based step is used to facilitate the predictions and preserve object edges. In their work, they develop novel online Expectation-Maximization (EM) methods for DCNN training under the weakly-supervised setting. Extensive experimental evaluation shows that the proposed techniques can learn models delivering competitive result from bounding boxes annotations. 

Scribbles have been widely used in connection with interactive image segmentation, being recognized as one of the most user-friendly ways for interacting. These approaches require the user to provide annotations interactively. The topic has been explored through graph cuts~\cite{boykov2001interactive}, random walks~\cite{grady2006random}, and weighted geodesic distances~\cite{bai2009geodesic}. As an improvement, \cite{tang2018regularized} proposes two regularization terms based on, respectively, normalized cuts and CRF. In this work, there are no extra inference steps explicitly generating masks, and their two loss terms are trained jointly with a partial cross-entropy loss function. In another work~\cite{kervadec2019constrained}, the authors enforce high-order (global) inequality constraints on the network output to leverage unlabelled data, guiding the training process with domain-specific knowledge (e.g. to constrain the size of the target region). To this end, they incorporate a differentiable penalty in the loss function avoiding expensive Lagrangian dual iterates and proposal generation. In the paper, the authors show a segmentation performance that is comparable to full supervision on three separate tasks.

Aiming at using scribbles to annotate images, ScribbleSup~\cite{lin2016scribblesup} learns CNN parameters by means of a graphical model that jointly propagates information from sparse scribbles to unlabelled pixels based on spatial constraints, appearance, and semantic content. In~\cite{tang2018normalized}, the authors propose a new principled loss function to evaluate the output of the network discriminating between labelled and unlabelled pixels, to avoid the poorer training that can result from standard loss functions, e.g. cross entropy, because of the presence of potentially mislabelled pixels in masks derived from scribbles or seeds. Unlike prior work, the cross entropy part of their loss evaluates only seeds where labels are known while a normalized cut term softly accounts for consistency of all pixels. Finally, \cite{xu2015learning} proposes a unified approach involving max-margin clustering (MMC) to take any form of weak supervision, e.g. tags, bounding boxes, and/or partial labels (strokes, scribbles), to infer pixel-level semantic labels, as well as learn an appearance model for each semantic class. Their loss function penalizes more or less the errors in positive or negative examples depending on whether the number of negative examples is larger than the positive examples or not.

In this paper, we focus on the use of scribbles as weak image annotations and propose a semantic segmentation approach that combines them with superpixels for propagating category information to obtain training masks, named as pseudo-masks because of the labelling mistakes they can contain. Further, we propose a specific loss function $L$ that makes use of those pseudo-masks, but at the same time intends to counteract their potential mistakes. To this end, $L$ consists of a partial cross-entropy term that uses the pseudo-masks, together with the Centroid Loss and a regularizing normalized MSE term that cooperate with the former to produce refined segmentation results through a joint training strategy. The Centroid Loss employs the labelling of pixels belonging to the scribbles to guide the training.

\section{Methodology}
\label{sc:methodology}

\begin{figure*}[t] % htb
    \centering
    \begin{tabular}{cc}
        \includegraphics[scale=0.3]{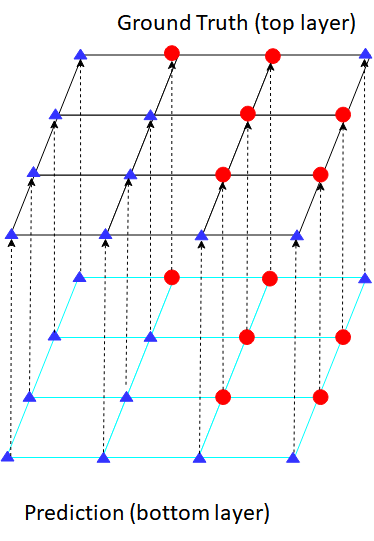} &
        \includegraphics[scale=0.3]{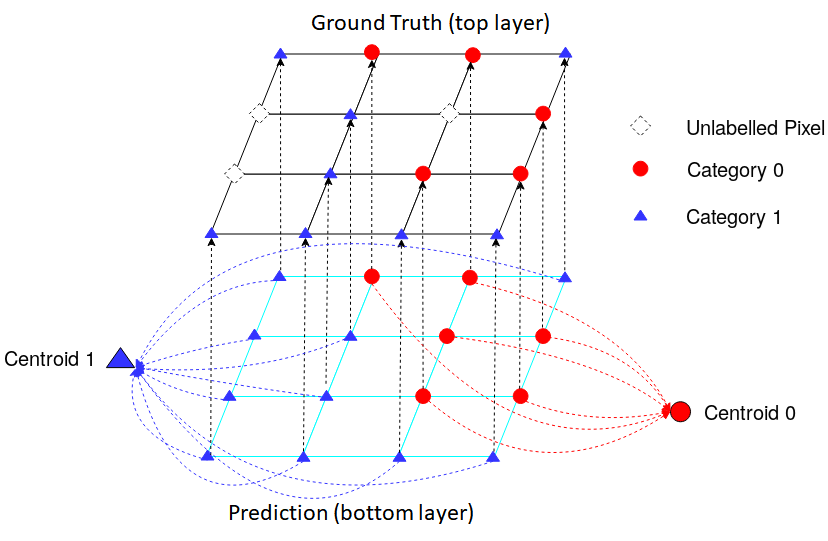} \\
        \footnotesize (a) & \footnotesize (b)
    \end{tabular}
    \caption{Illustration of (a) full supervision and (b) our weakly-supervised approach for semantic segmentation: (a) all pixels are labelled to make the prediction [bottom layer of the drawing] resemble the ground truth [top layer of the drawing] as much as possible after pixel-wise training; (b) to solve the WSSS problem, the category information from the incomplete ground truth, i.e. the weak annotations, is propagated towards the rest of pixels making use of pixel similarity and minimizing distances to class centroids derived from the weak annotations. }
    \label{fig:cluster_segmentation}
\end{figure*}

Figure~\ref{fig:cluster_segmentation}(a) illustrates fully supervised semantic segmentation approaches based on DCNN, which, applying a pixel-wise training strategy, try to make network predictions resemble the full labelling as much as possible, thus achieving good segmentation performance levels in general. By design, this kind of approach ignores the fact that pixels of the same category tend to be similar to their adjacent pixels. This similarity can, however, be exploited when addressing the WSSS problem by propagating the known pixel categories towards unlabelled pixels. In this respect, several works reliant on pixel-similarity to train the WSSS network can be found in the literature: e.g. a dense CRF is used in~\cite{papandreou2015weakly}, the GraphCut approach is adopted in~\cite{zhao2018pseudo}, and superpixels are used in ScribbleSup~\cite{lin2016scribblesup}. 

Inspired by the aforementioned, in this work, we propose a semantic segmentation approach using scribble annotations and a specific loss function intended to compensate for missing labels and errors in the training masks. To this end, class centroids determined from pixels coinciding with the scribbles, whose labelling is actually the ground truth of the problem, are used in the loss function to guide the training of the network so as to obtain improved segmentation outcomes. The process is illustrated in Fig.~\ref{fig:cluster_segmentation}(b).

\begin{figure*}[t] % htb
    \centering
    \begin{tabular}{@{\hspace{0mm}}c@{\hspace{1mm}}c@{\hspace{1mm}}c@{\hspace{1mm}}c@{\hspace{0mm}}}
        \includegraphics[width=35mm,height=35mm]{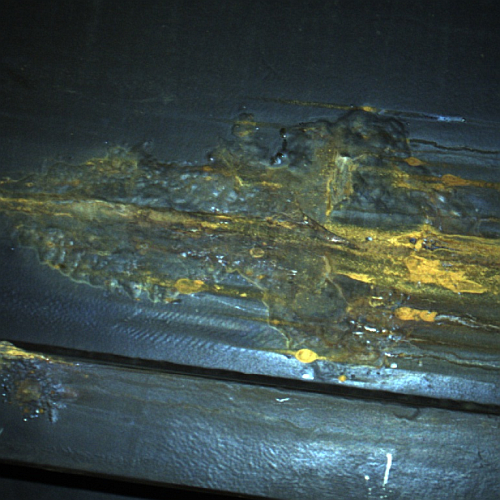}
        &
        \includegraphics[width=35mm,height=35mm]{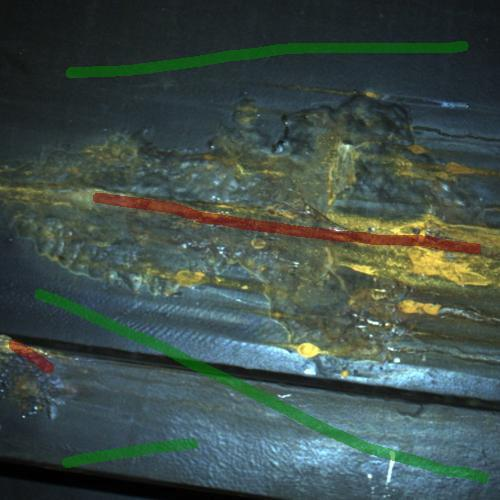}
        &
        \includegraphics[width=35mm,height=35mm]{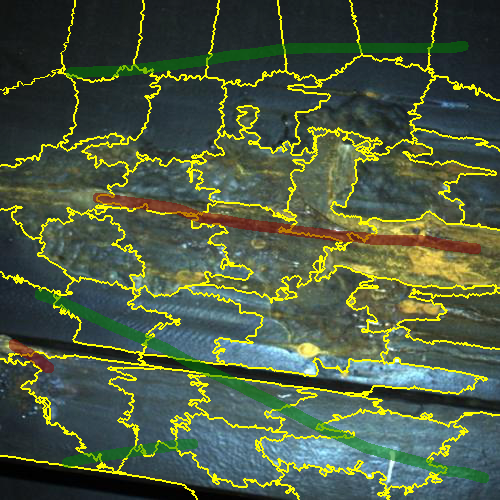}
        &
        \includegraphics[width=35mm,height=35mm]{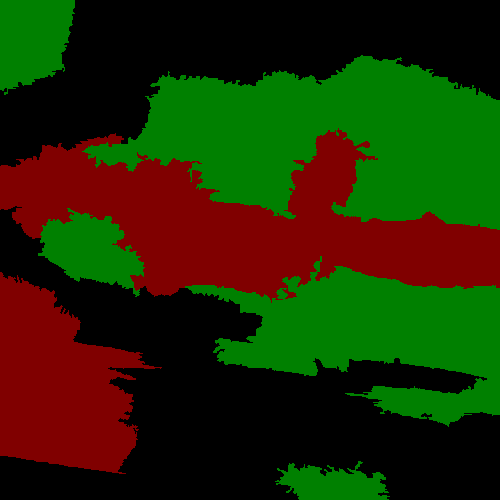} 
        \\
        \includegraphics[width=35mm,height=35mm]{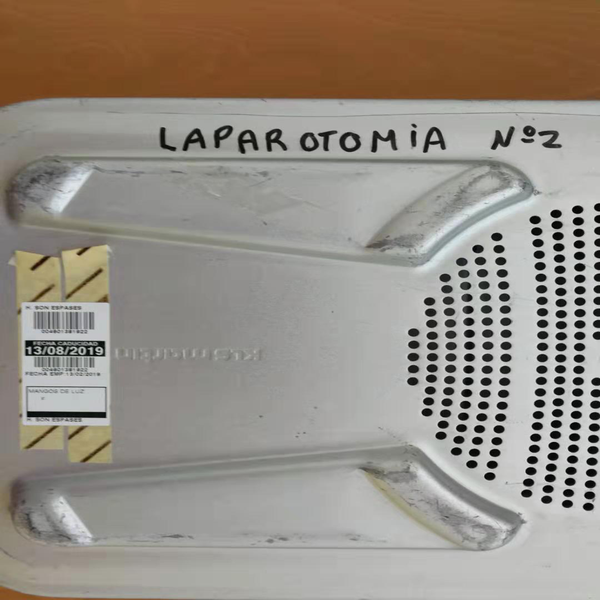}
        &
        \includegraphics[width=35mm,height=35mm]{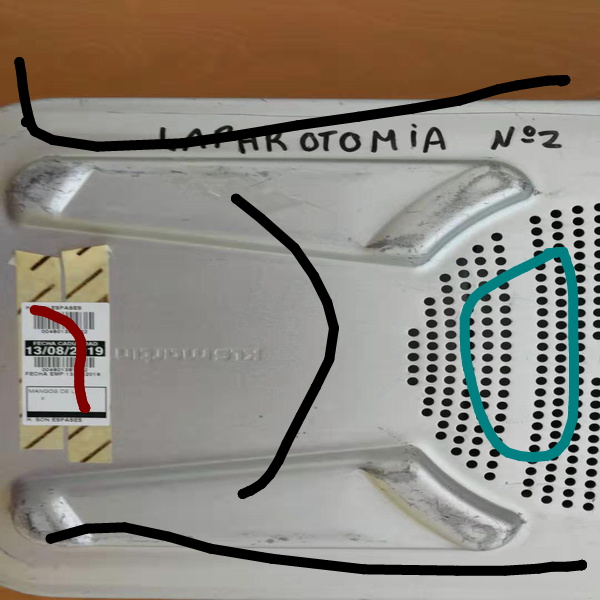}
        &
        \includegraphics[width=35mm,height=35mm]{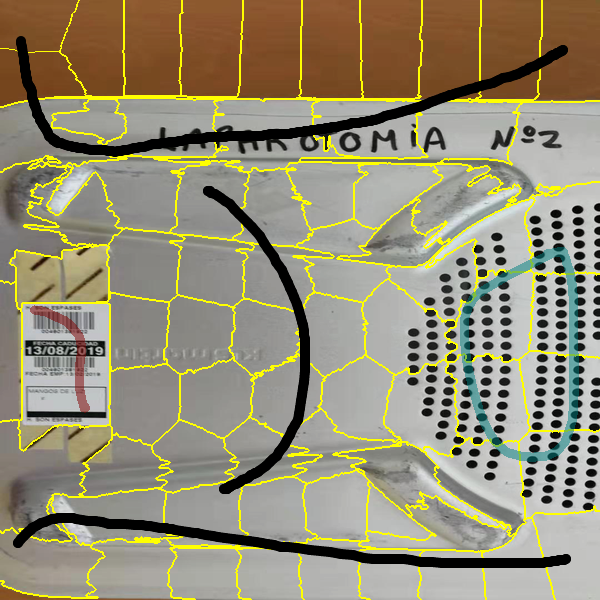}
        &
        \includegraphics[width=35mm,height=35mm]{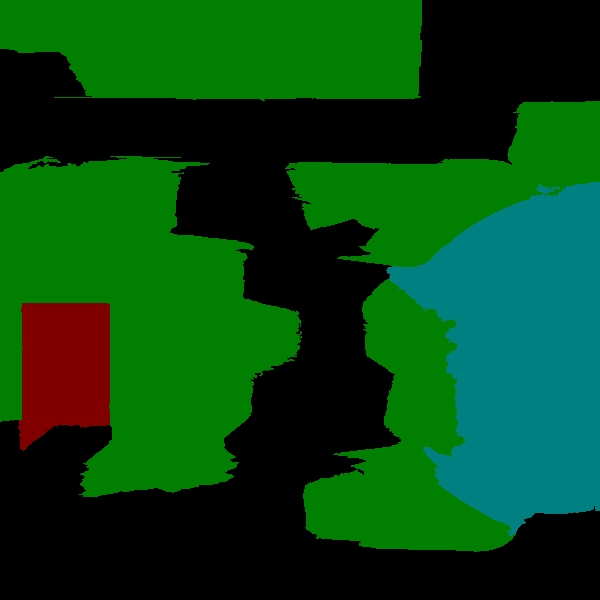} \\ 
        \footnotesize (a) & \footnotesize (b) & \footnotesize (c) & \footnotesize (d)
    \end{tabular}
    %\vspace{-2mm}
    \caption{Weak annotation and propagation example: (a) original images; (b) scribbles superimposed over the original image; (c) scribbles superimposed over the superpixels segmentation result; (d) resulting pseudo-masks. Regarding the scribble annotations: (1st row) red and green scribbles respectively denote corrosion and background; (2nd row) black, red and blue scribbles respectively denote background, tracking label and the internal filter texture. As for the pseudo-masks: (1st row) red, black and green pixels respectively denote corrosion, background and unlabelled pixels; (2nd row) red, blue, black and green pixels respectively denote the tracking label, the internal filter texture, the background and the unlabelled pixels.}
    \label{fig:weak_labels}
\end{figure*}

Furthermore, similarly to ScribbleSup~\cite{lin2016scribblesup}, we also combine superpixels and scribble annotations to propagate category information and generate pseudo-masks as segmentation proposals, thus making the network converge fast and achieve competitive performance. By way of example, Fig.~\ref{fig:weak_labels}(b) and (c) show, respectively, the scribble annotations and the superpixels-based segmentations obtained for two images of the two application cases considered. The corresponding pseudo-masks, containing more annotated pixels than the scribbles, are shown in Fig.~\ref{fig:weak_labels}(d). As can be observed, not all pixels of the pseudo-masks are correctly labelled, what may affect segmentation performance. It is because of this fact that we incorporate the Centroid Loss and a normalized MSE term into the full loss function. This is discussed in Section~\ref{sec:cen_loss}.

The remaining methodological details are given along the rest of this section: we begin with the way how weak annotations are handled and how the pseudo-masks are obtained in Section~\ref{sec:pseudo_mask}, while the architecture of the network is described in Section~\ref{sec:net_arch} and the different loss terms are detailed and discussed in Sections~\ref{sec:pce_loss} (partial Cross-Entropy loss, $L_\text{pCE}$), \ref{sec:cen_loss} (Centroid Loss, $L_\text{cen}$) and~\ref{sec:full_loss} (normalized MSE-term, $L_\text{mse}$, and the full loss function $L$).

\subsection{Weak annotations and pseudo-masks generation}
\label{sec:pseudo_mask}

As already said, Fig.~\ref{fig:weak_labels}(b) shows two examples of scribble annotations, one for the visual inspection case (top) and the other for the quality control case (bottom). Because scribbles represent only a few pixels, the segmentation performance that the network can be expected to achieve will be far from satisfactory for any task that is considered. To enhance the network performance, we combine the scribbles with an oversegmentation of the image to generate pseudo-masks as segmentation proposals for training. For the oversegmentation, we make use of the Adaptive-SLIC (SLICO) algorithm~\cite{achanta2012slic}, requesting enough superpixels so as not to mix different classes in the same superpixel. Figure~\ref{fig:weak_labels}(c,top) shows an oversegmentation in 50 superpixels, while 80 are selected for Fig.~\ref{fig:weak_labels}(c,bottom). Next, those pixels belonging to a superpixel that intersects with a scribble are labelled with the same class as the scribble, as shown in Fig.~\ref{fig:weak_labels}(d). In Fig.~\ref{fig:weak_labels}(d,top), the black pixels represent the background, the red pixels indicate corrosion, and the green pixels denote unlabelled pixels. In Fig.~\ref{fig:weak_labels}(d,bottom), black and green pixels denote the same as for the top mask, while the red pixels represent the tracking label and the blue pixels refer to the internal filter class.

\subsection{Network Architecture}
\label{sec:net_arch}

In this work, we adopt U-Net~\cite{Ronneberger2015} as the base network architecture. As it is well known, U-Net evolves from the fully convolutional neural network concept and consists of a contracting path followed by an expansive path. It was developed for biomedical image segmentation, though it has been shown to exhibit good performance in general for natural images even for small training sets. Furthermore, we also embed Attention Gates (AG) in U-Net, similarly to Attention U-Net (AUN)~\cite{oktay2018attention}. These attention modules have been widely used in e.g. Natural Language Processing (NLP)~\cite{vaswani2017attention,clark2019does,serrano2019attention,jain2019attention}. Other works related with image segmentation~\cite{hu2018squeeze,jetley2018learn,oktay2018attention,sinha2020multi} have introduced them for enhanced performance. In our case, AGs are integrated into the decoding part of U-Net to improve its ability to segment small targets. 

\begin{figure*}%[t] % htb
    \centering
    \includegraphics[scale=0.35]{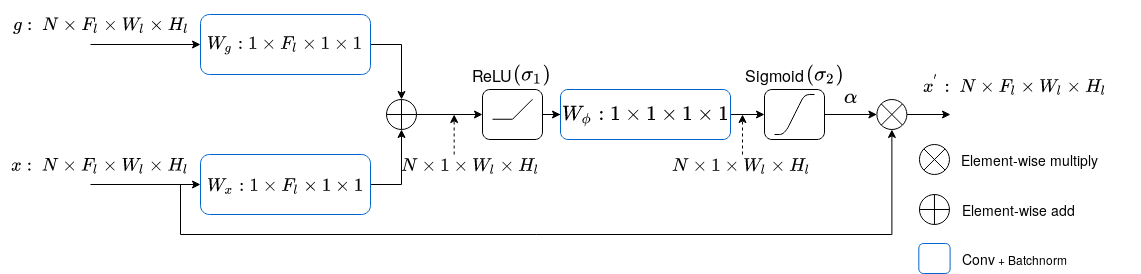}
    \caption{Schematic diagram of an Attention Gate (AG). $N$ is the size of the mini-batch.}
    \label{fig:AG}
\end{figure*}

For completeness, we include in Fig.~\ref{fig:AG} a schematic about the operation of the AG that we make use of in this work, which, in our case, implements (\ref{eq:AG}) as described below: 
\begin{align}
    (x_{i,c}^l)^\prime &= \alpha_i^l \, x_{i,c}^l \label{eq:AG} \\
    \alpha_i^l &= \sigma_2(W_{\phi}^T(\sigma_1(W_x^T x_i^l + W_g^T g_i + b_g)) + b_{\phi}) \nonumber 
\end{align}
where the feature-map $x_i^l \in \mathbb{R}^{F_l}$ is obtained at the output of layer $l$ for pixel $i$, $c$ denotes a channel in $x_{i,c}^l$, $F_l$ is the number of feature maps at that layer, the gating vector $g_i$ is used for each pixel $i$ to determine focus regions and is such that $g_i \in \mathbb{R}^{F_l}$ (after up-sampling the input from the lower layer), $W_g \in \mathbb{R}^{F_l \times 1}$, $W_x \in \mathbb{R}^{F_l \times 1}$, and $W_{\phi} \in \mathbb{R}^{1 \times 1}$ are linear mappings, while $b_g \in \mathbb{R}$ and $b_{\phi} \in \mathbb{R}$ denote bias terms, $\sigma_1$ and $\sigma_2$ respectively represent the ReLU and the sigmoid activation functions, $\alpha_i^l \in [0,1]$ are the resulting attention coefficients, and $\Phi_\text{att} = \{W_g, W_x, b_g; W_{\phi}, b_{\phi}\}$ is the set of parameters of the AG. 

The attention coefficients $\alpha_i$ are intended to identify salient image regions and discard feature responses so as to preserve only the activations relevant to the specific task. In~\cite{hu2018squeeze}, the Squeeze-and-Excitation (SE) block obtains attention weights in channels for filter selection. In our approach, the AGs involved calculate attention weights at the spatial level. 

\begin{figure*}%[t] % htb
    \centering
    \includegraphics[scale=0.25]{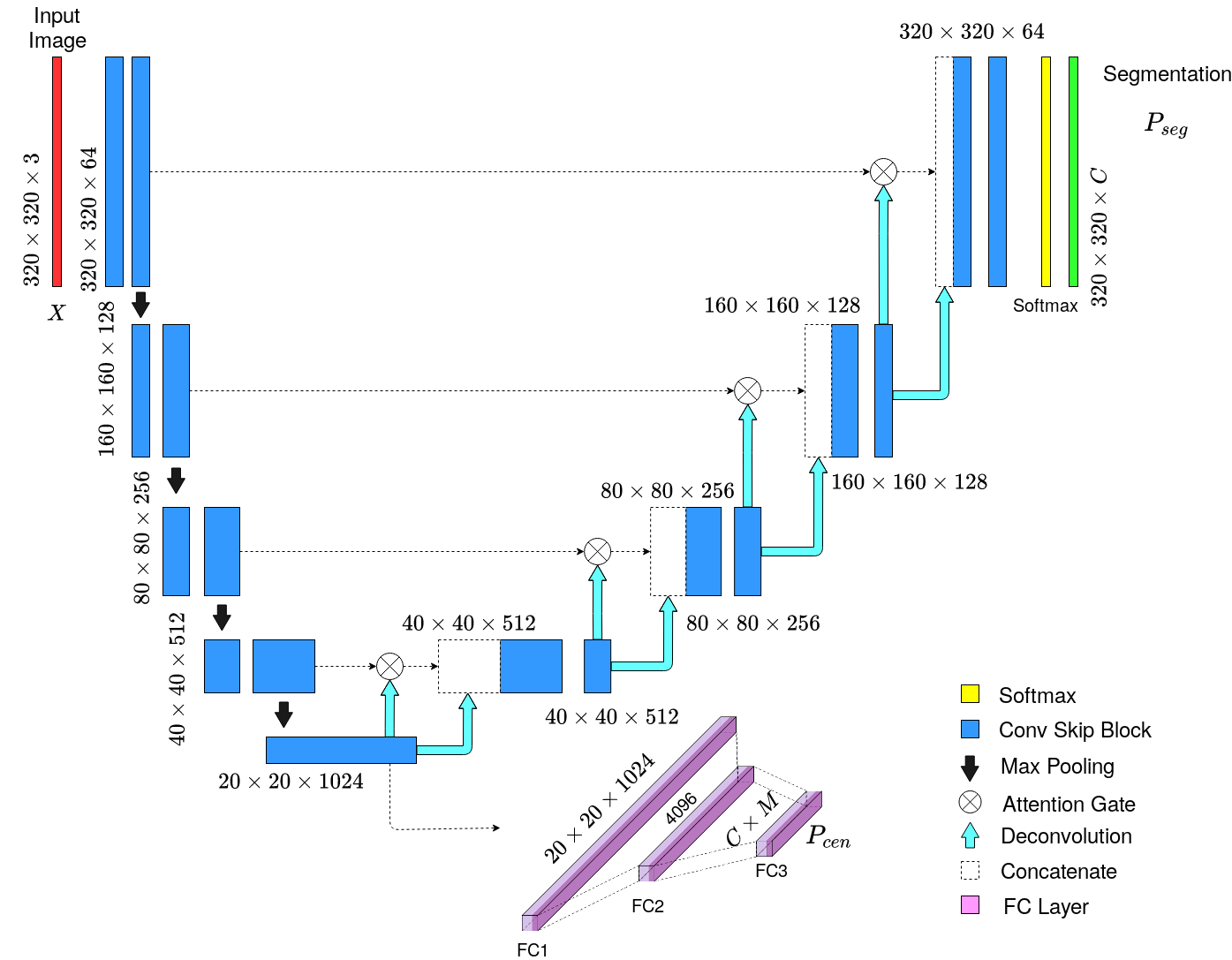}
    \caption{Block diagram of the Centroids AUN model. The size decreases gradually by a factor of 2 at each scale in the encoding part and increases by the same factor in the decoding part. In the latter, AGs are used to help the network focus on the areas of high-response in the feature maps. The \textit{Conv~Skip} block is the \textit{skip~connection} of ResNet \cite{he2016deep}. The sub-network of the lower part of the diagram is intended to predict class centroids. In the drawing, $C$ denotes the number of classes and $M$ is the dimension of the class centroids.}
    \label{fig:Unet_cluster}
\end{figure*}

As shown in Fig.~\ref{fig:Unet_cluster}, AGs are fed by two input tensors, one from the encoder side of U-Net and the other from the decoder side, respectively $x$ and $g$ in Fig.~\ref{fig:AG}. With the AG approach, spatial regions are selected on the basis of both the activations $x$ and the contextual information provided by the gating signal $g$ which is collected from a coarser scale. The contextual information carried by the gating vector $g$ is hence used to highlight salient features that are passed through the skip connections. In our case, $g$ enters the AG after an up-sampling operation that makes $g$ and $x$ have compatible shapes (see Fig.~\ref{fig:AG}).

Apart from the particularities of the AG that we use, which have been described above, another difference with the original AUN is the sub-network that we attach to the main segmentation network, as can be observed from the network architecture that is shown in Fig.~\ref{fig:Unet_cluster}. This sub-network is intended to predict class centroids on the basis of the scribbles that are available in the image, with the aim of improving the training of the main network from the possibly noisy pseudo-masks, and hence achieve a higher level of segmentation performance. Consequently, during training: (1) our network handles two sorts of ground truth, namely scribble annotations $Y_\text{scr}$ to train the attached sub-network for proper centroid predictions, and the pseudo-masks $Y_\text{seg}$ for image segmentation; and (2) the augmented network yields two outputs, a set of centroids $P_\text{cen}$ and the segmentation of the image $P_\text{seg}$ (while during inference only the segmentation output $P_\text{seg}$ is relevant). Predicted cluster centroids are used to calculate the Centroid Loss term $L_\text{cen}$ (described in Section~\ref{sec:cen_loss}) of the full loss function $L$, which comprises two more terms (as described in Section~\ref{sec:full_loss}). Thanks to the design of $L$, the full network --i.e. the AUN for semantic segmentation and the sub-net for centroids prediction-- is trained through a joint training strategy following an end-to-end learning model. During training, the optimization of $L_\text{cen}$ induces updates in the main network weights via back-propagation that are intended to reach enhanced training and therefore produce better segmentations. 

As can be observed, the centroids prediction sub-net is embedded into the intermediate part of the network, being fed by the last layer of the encoder side of our AUN. As shown in Fig.~\ref{fig:Unet_cluster}, this sub-net consists of three blocks, each of which comprises a fully connected layer, a batch-normalization layer, and a ReLU activation function. The shape of $P_\text{cen}$ is $C\times M$, where $C$ is the number of categories and $M$ denotes the dimension of the feature space where the class centroids are defined. In our approach, centroid features are defined from the softmax layer of the AUN, and hence comprises $C$ components, though we foresee to combine them with $K$ additional features from the classes which are incorporated externally to the operation of the network, and hence $M = C+K$. On the other side, the shape of $P_\text{seg}$ is $C\times W\times H$, where $(H,W)$ is the size of the input image.

\subsection{Partial Cross-Entropy Loss}
\label{sec:pce_loss}

Given a $C$-class problem and a training set $\Omega$, comprising a subset $\Omega_L$ of labelled pixels and a subset $\Omega_U$ of unlabelled pixels, the Partial Cross-Entropy Loss $L_\text{pCE}$, widely used for WSSS, computes the cross-entropy only for labelled pixels $p \in \Omega_L$, ignoring $p \in \Omega_U$:
\begin{equation}
    L_\text{pCE} = \sum\limits_{c=1}^{C}\sum\limits_{p\in \Omega_{L}^{(1)}} -y_{g(p),c}~\log~y_{s(p),c}
    \label{func:partial_cross-entropy}
\end{equation}
where $y_{g(p),c} \in \{0,1\}$ and $y_{s(p),c} \in [0,1]$ represent respectively the ground truth and the segmentation output. In our case, and for $L_\text{pCE}$, $\Omega_L^{(1)}$ is defined as the pixels labelled in the pseudo-masks (hence, pixels from superpixels not intersecting with any scribble belong to $\Omega_U$ and are not used by (\ref{func:partial_cross-entropy})). Hence, $y_{g(p),c}$ refers to the pseudo-masks, i.e. $Y_\text{seg}$, while $y_{s(p),c}$ is the prediction, i.e. $P_\text{seg}$, as supplied by the softmax final network layer. 

% In this way, $L_\text{pCE}$ exploits the similarity between adjacent pixels. , i.e. $Y_\text{seg}$ , i.e. $P_\text{seg}$

\subsection{Centroid Loss}
\label{sec:cen_loss}

As can be easily foreseen, when the network is trained using the pseudo-masks, the segmentation performance depends on how accurate the pseudo-masks are and hence on the quality of superpixels, i.e. how they adhere to object boundaries and avoid mixing classes. The Centroid Loss function is introduced in this section for the purpose of compensating a dependence of this kind and improving the quality of the segmentation output. 

In more detail, we define the Centroid Loss term $L_\text{cen}$ as another partial cross-entropy loss:
\begin{equation}
    L_\text{cen} = \sum\limits_{c=1}^{C} \sum\limits_{p\in \Omega_{L}^{(2)}} -y_{g(p),c}^{*}~\log~y_{s(p),c}^{*} 
    \label{func:cen_loss}
\end{equation}
defining in this case: 
\begin{itemize}
\item $\Omega_{L}^{(2)}$ as the set of pixels coinciding with the scribbles, 
\item $y_{g(p),c}^{*}$ as the corresponding labelling, and
\end{itemize}
\begin{align}
    y_{s(p),c}^{*} &= \frac{\exp(-d_{p,c})}{\sum\limits^{C}_{c^{'}=1} \exp(-d_{p,c^{'}})} \label{func:cen_loss_2} \\
    d_{p,c} &=\frac{||f_p-\mu_c||_2^2}{\sum\limits^{C}_{c^{'}=1}||f_p-\mu_{c^{\prime}}||_2^2} \label{func:cen_loss_3}
\end{align}
where: (1) $f_p$ is the feature vector associated to pixel $p$ and (2) $\mu_c$ denotes the centroid predicted for class $c$, i.e. $\mu_c \in P_\text{cen}$. $f_p$ is built from the section of the softmax layer of the main network corresponding to pixel $p$, though $f_p$ can be extended with the incorporation of additional external features, as already mentioned. This link between $L_\text{pCE}$ and $L_\text{cen}$ through the softmax layer makes both terms decrease through the joint optimization, in the sense that for a reduction in $L_\text{cen}$ to take place, and hence in the full loss $L$, also $L_\text{pCE}$ has to decrease by better predicting the class of the pixels involved. The additional features that can be incorporated in $f_p$ try to introduce information from the classes, e.g. predominant colour, to guide even more the optimization.

In practice, this loss term \textit{pushes} pixel class predictions towards, ideally, a subset of the corners of the C-dimensional hypercube, in accordance with the scribbles, i.e. the available ground truth. Some similarity can be certainly established with the K-means algorithm. Briefly speaking, K-means iteratively calculates a set of centroids for the considered number of clusters/classes, and associates the samples to the closest cluster in feature space, thus minimizing the intra-class variance until convergence. Some DCNN-based clustering approaches reformulate K-means as a neural network optimizing the intra-class variance loss by means of a back-propagation-style scheme~\cite{Wen2016,Peng2019}. Differently from the latter, in this work, (\ref{func:cen_loss}) reformulates the unsupervised process of minimizing the distances from samples to centroids into a supervised process since the clustering takes place around the true classes defined by the labelling of the scribbles $y_{g(p),c}^{*}$ and the extra information that may be incorporated.

\subsection{Full Loss Function}
\label{sec:full_loss}

Since $L_\text{pCE}$ applies only to pixels labelled in the pseudo-mask and $L_\text{cen}$ is also restricted to a subset of image pixels, namely the pixels coinciding with the scribbles, we add a third loss term in the form of a normalized MSE loss $L_\text{mse}$ to behave as a regularization term that involves all pixels for which a class label must be predicted $\Omega_{L}^{(3)}$, i.e. the full image. This term calculates the normalized distances between the segmentation result for every pixel and its corresponding centroid:
\begin{equation}
    %L_\text{mse} = \frac{\sum\limits_{c=1}^{C} \sum\limits_{p\in \Omega_{L}^{(3)}} d_{p,c}}{C \cdot |\Omega_{L}^{(3)}|}
    %L_\text{mse} = \frac{\sum\limits_{p\in \Omega_{L}^{(3)}} \displaystyle\min_{c = 1\,..\,C}\{d_{p,c}\}}{|\Omega_{L}^{(3)}|}
    L_\text{mse} = \frac{\sum\limits_{p\in \Omega_{L}^{(3)}} d_{p,c(p)}}{|\Omega_{L}^{(3)}|}
    \label{func:mse_reg}
\end{equation}
where $|\mathcal{A}|$ stands for the cardinality of set $\mathcal{A}$, and $d_{p,c(p)}$ is as defined by (\ref{func:cen_loss_3}), with $c(p)$ as the class prediction for pixel $p$ (and $\mu_{c(p)}$ the corresponding predicted centroid), taken from the softmax layer.
%where $|\mathcal{A}|$ stands for the cardinality of set $\mathcal{A}$, and $d_{p,c}$ is as defined by (\ref{func:cen_loss_3}).

Finally, the complete loss function is given by:
\begin{equation}
    L = L_\text{pCE} + \lambda_\text{cen} L_\text{cen} + \lambda_\text{mse} L_\text{mse}
    \label{func:final_loss}
\end{equation}
where $\lambda_\text{cen}$ and $\lambda_\text{mse}$ are trade-off constants.

\section{Experiments and Discussion}
\label{sc:experiments}

In this section, we report on the results obtained for the two application cases that constitute our benchmark. For a start, Section~\ref{sec:exp_setup} describes the experimental setup. Next, in Section~\ref{sec:dis_feature}, we discuss about the feature space where the Centroid Loss is defined and its relationship with the weak annotations, while Section~\ref{sec:effect_cen} evaluates the effect on the segmentation performance of several combinations of the terms of the loss function $L$, and Section~\ref{sec:influence_weak} analyzes the impact of weak annotations and their propagation. Subsequently, our approach is compared against two previously proposed methods in Section~\ref{sec:performance_compare}. To finish, we address final tuning and show segmentation results, for qualitative evaluation purposes, for some images of both application cases in Section~\ref{sec:result_displays}.

\subsection{Experimental Setup}
\label{sec:exp_setup}

\subsubsection{Datasets}

The dataset from the quality control application case consists of a total of 484 images, two thirds of which are designated for training and the rest for testing. Regarding the dataset for the visual inspection application, it comprises 241 images and the same strategy is adopted for splitting into the training and test sets. Both datasets have been in turn augmented with rotations and scaled versions of the original images, together with random croppings, to increase the diversity of the training set. Finally, as already explained, the ground truth for both datasets comprises scribbles and pseudo-masks (generated in accordance to the process described in Section~\ref{sec:pseudo_mask}). 

\begin{figure}[t] % htb
    \centering
    \begin{tabular}{@{\hspace{-3mm}}c@{\hspace{1mm}}c@{\hspace{1mm}}c@{\hspace{1mm}}c@{\hspace{0mm}}}
        \includegraphics[width=0.23\columnwidth,height=0.23\columnwidth]{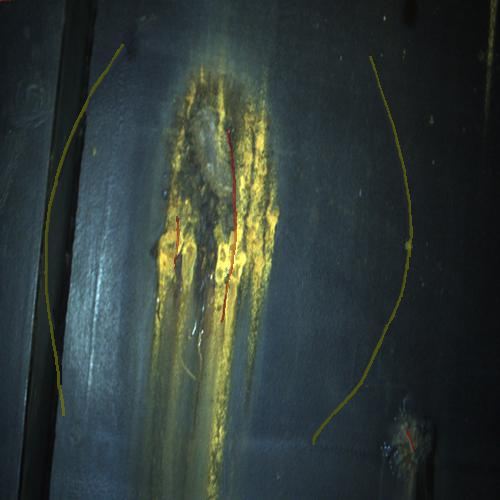}
        &
        \includegraphics[width=0.23\columnwidth,height=0.23\columnwidth]{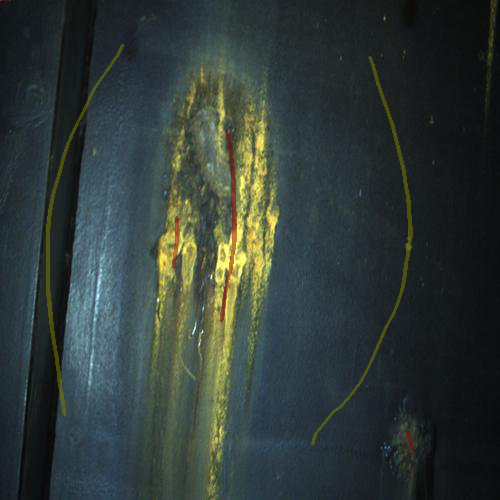}
        &
        \includegraphics[width=0.23\columnwidth,height=0.23\columnwidth]{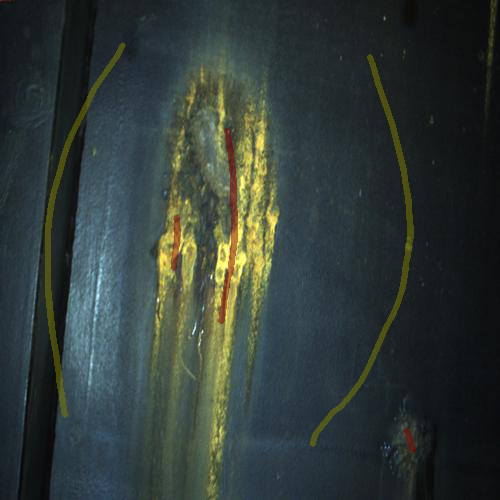}
        &
        \includegraphics[width=0.23\columnwidth,height=0.23\columnwidth]{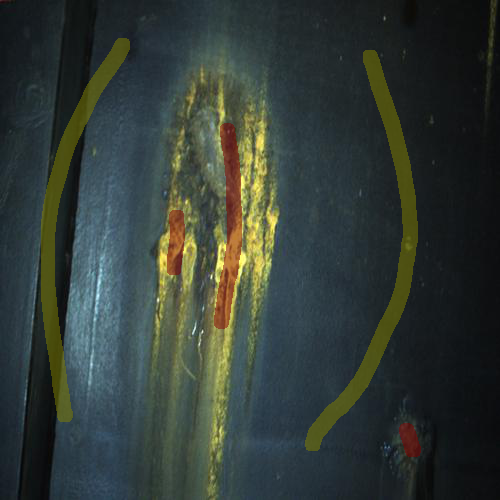} \\[-1mm]
        \footnotesize 2 & \footnotesize 5 & \footnotesize 10 & \footnotesize 20 \\
        \includegraphics[width=0.23\columnwidth,height=0.23\columnwidth]{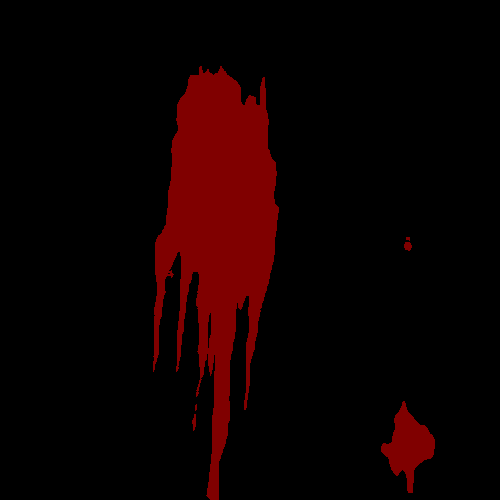}
        &
        \includegraphics[width=0.23\columnwidth,height=0.23\columnwidth]{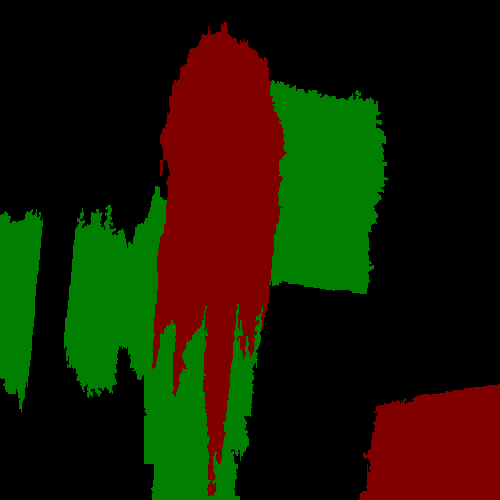}
        &
        \includegraphics[width=0.23\columnwidth,height=0.23\columnwidth]{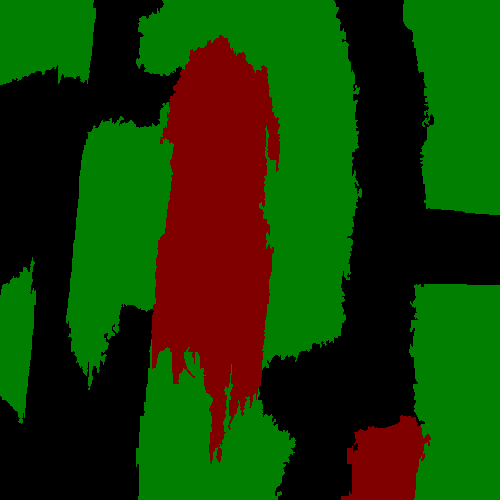}
        &
        \includegraphics[width=0.23\columnwidth,height=0.23\columnwidth]{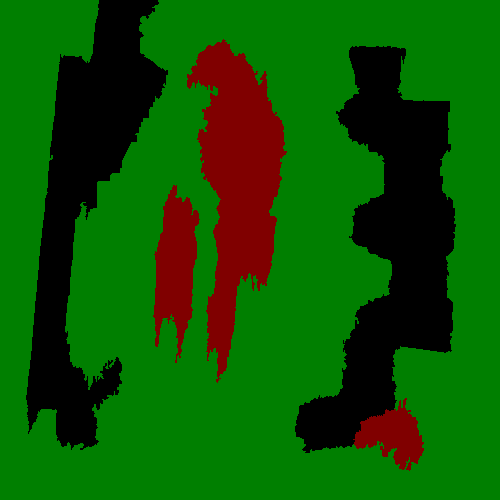} \\[-1mm]
        \footnotesize full mask & \footnotesize 30 & \footnotesize 50 & \footnotesize 80 \\
        \includegraphics[width=0.23\columnwidth,height=0.23\columnwidth]{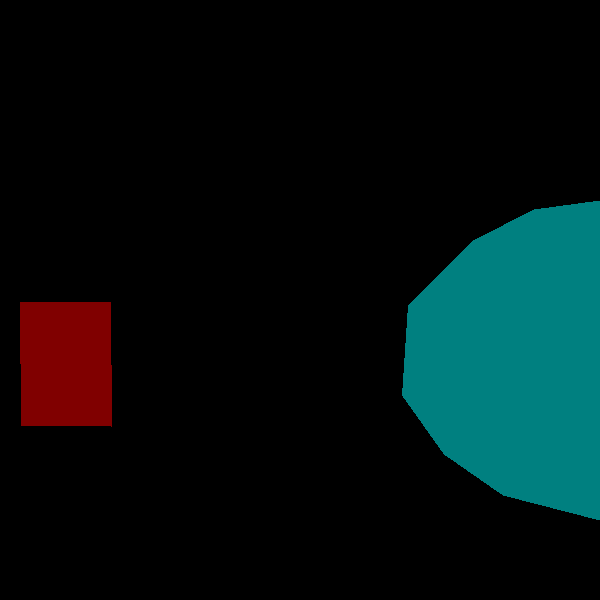}
        &
        \includegraphics[width=0.23\columnwidth,height=0.23\columnwidth]{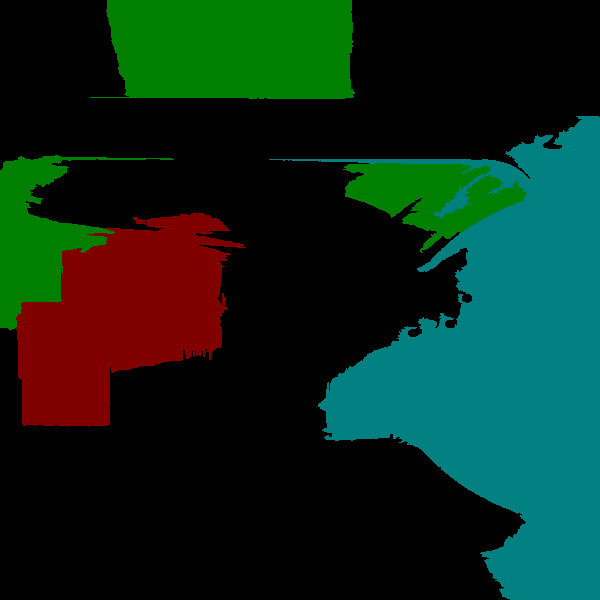}
        &
        \includegraphics[width=0.23\columnwidth,height=0.23\columnwidth]{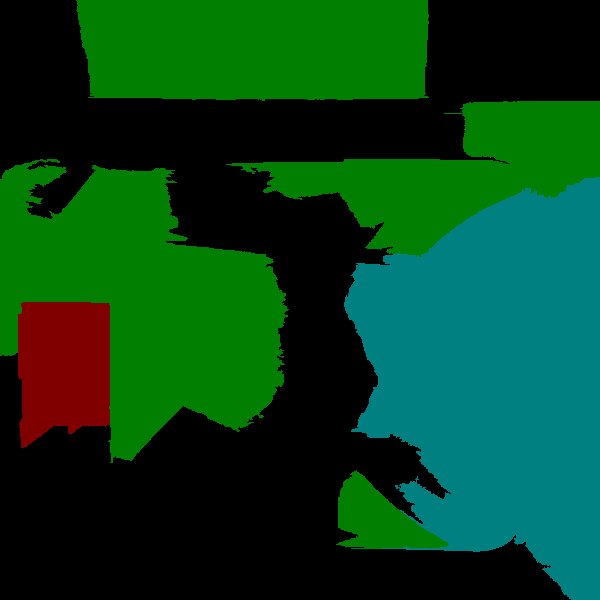}
        &
        \includegraphics[width=0.23\columnwidth,height=0.23\columnwidth]{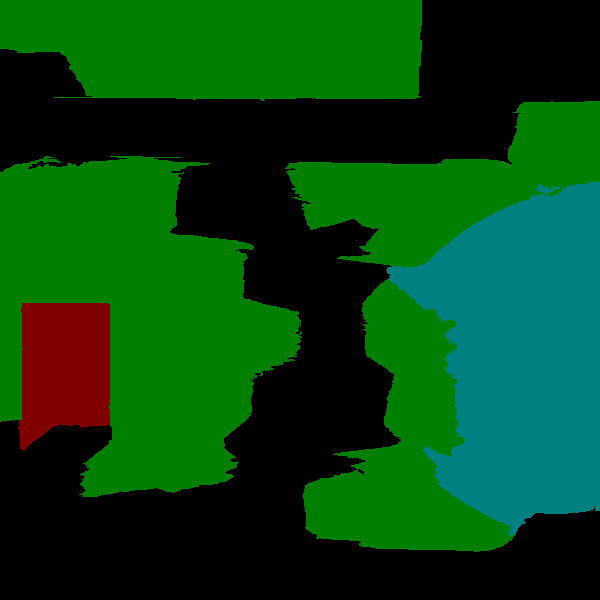} \\[-1mm]
        \footnotesize full mask & \footnotesize 30 & \footnotesize 50 & \footnotesize 80 
    \end{tabular}
    %\vspace{-2mm}
    \caption{Examples of weak annotations and their propagation for the two application cases: (1st row) examples of scribble annotations of different widths, namely, from left to right, 2, 5, 10 and 20 pixels, for the visual inspection case; (2nd and 3rd rows) the leftmost image shows the fully supervised ground truth, while the remaining images are examples of pseudo-masks generated from 20-pixel scribbles and for different amounts of superpixels, namely 30, 50, and 80, for the two images of Fig.~\ref{fig:weak_labels} and, hence, for the visual inspection and the quality control application cases. (The colour code is the same as for Fig.~\ref{fig:weak_labels}.)}
    \label{fig:pseudo_mask_exp}
\end{figure}

By way of illustration, Fig.~\ref{fig:pseudo_mask_exp} shows, for the two application cases, some examples of weak annotations with different settings as for the width of the scribbles and the number of superpixels used for generating the pseudo-masks.

\subsubsection{Evaluation metrics}
For quantitative evaluation of our approach, we consider the following metrics:
\begin{itemize}

\item The mean Intersection Over Union (mIOU), which can be formally stated as follows: given $n_{ij}$ as the number of pixels of class $i$ that fall into class $j$, for a total of $C$ different classes, the mIOU is defined as (\ref{func:miou}), see e.g.~\cite{long2015fully}, 
\begin{equation}
    \text{mIOU} = \frac{1}{C} \sum_i \frac{n_{ii}}{\sum_j n_{ij} + \sum_j n_{ji} - n_{ii}}\,.
    \label{func:miou}
\end{equation}
% In order to precisely represent the quality of weak annotations, we also compute a weak mIOU (wmIOU) based on (\ref{func:miou}) using the weak annotations as ground truth, i.e. the scribbles and their propagation in the form of pseudo-masks by means of superpixels.

\item The mean Recall and mean Precision are also calculated to evaluate the segmentation performance for all classes. True Positive (TP), False Positive (FP) and False Negative (FN) samples are determined from the segmentation results and the ground truth. Using a macro-averaging
approach~\cite{Zhang2014}, the mean Recall (mRec) and mean Precision (mPrec) are expressed as follows:
    \begin{align}
        \text{mRec}  & = \frac{1}{C} \left( \sum_i \frac{TP_i}{TP_i + FN_i} \right) = \frac{1}{C} \left( \sum_i \frac{TP_i}{T_i} \right) 
    \label{func:rec_prec_1}
        \\
        \text{mPrec} & = \frac{1}{C} \left( \sum_i \frac{TP_i}{TP_i + FP_i} \right) = \frac{1}{C} \left( \sum_i \frac{TP_i}{P_i} \right)
    \label{func:rec_prec_2}
    \end{align}
where $TP_i$, $FP_i$ and $FN_i$ are, respectively, the true positives, false positives and false negatives for class $i$, and $T_i$ and $P_i$ are, respectively, the number of positives in the ground truth and the number of predicted positives, both for class $i$. From now on, to shorten the notation, when we refer to precision and recall, it must be understood that we are actually referring to mean precision and mean recall.
% From now on, we will refer to simply precision and recall, i.e. we will drop the \textit{mean} prefix, to shorten the notation. 
\item The F$_1$ score as the harmonic mean of precision and recall:
\begin{equation}
    F_1 = \frac{2\cdot\text{mPrec}\cdot\text{mRec}}{\text{mPrec} + \text{mRec}}
\end{equation} 
\end{itemize}

In all experiments, we make use of fully supervised masks/ground truth for both datasets in order to be able to report accurate calculations about the segmentation performance. This ground truth has been manually generated only for this purpose, it has been used for training only when referring to the performance of the full-supervised approach, for comparison purposes between the full- and weakly-supervised solutions.

To finish, in a number of experiments we also report on the quality of the pseudo-masks, so that the segmentation performance reported can be correctly valued. To this end, we calculate a weak mIOU (wmIOU) using \ref{func:miou} between the psedo-mask and the fully-supervised mask involved.

%\subsubsection{Details of implementation}
\subsubsection{Implementation details and main settings}

All experiments have been conducted using the Pytorch framework running in a PC fitted with an NVIDIA GeForce RTX 2080 Ti GPU, a 2.9GHz 12-core CPU with 32 GB RAM, and Ubuntu 64-bit. The batch size is 8 for all experiments and the size of the input image is $320\times 320$ pixels, since this has turned out to be the best configuration for the aforementioned GPU. 

As already mentioned, the AUN for semantic segmentation and the sub-net for centroid prediction are jointly trained following an end-to-end learning model. The network weights are initialized by means of the Kaiming method~\cite{He2015b}, and they are updated using a $10^{-4}$ learning rate for 200 epochs. 

Best results have been obtained for the balance parameters $\lambda_\text{cen}$ and $\lambda_\text{mse}$ set to 1. 

%%\subsubsection{The Setting of Comparative Experiments} 
\subsubsection{Overall view of the experiments} 

The experiments that are going to be discussed along the next sections consider different configurations for the different elements that are involved in our semantic segmentation approach. These configurations, which are enumerated in Table~\ref{tab:exp_define}, involve: 
\begin{itemize}
\item different widths of the scribble annotations used as ground truth, namely 2, 5, 10 and 20 pixels, 
\item different amounts of superpixels for generating the pseudo-masks, namely 30, 50 and 80, 
\item two ways of defining the feature space for the class centroids: from exclusively the softmax layer of AUN and combining those features with other features from the classes.
\end{itemize}
Notice that the first rows of Table~\ref{tab:exp_define} refer to experiments where the loss function used for training is just the partial cross-entropy, as described in (\ref{func:partial_cross-entropy}), and therefore can be taken as a lower baseline method. The upper baseline would correspond to the configuration using full masks and the cross entropy loss $L_\text{CE}$ for training, i.e. full supervised semantic segmentation, which can also be found in Table~\ref{tab:exp_define} as the last row.

Apart from the aforementioned variations, we also analyse the effect of several combinations of the loss function terms, as described in (\ref{func:final_loss}), defining three groups of experiments: Group 1 (G1), which indicates that the network is trained by means of only $L_\text{pCE}$, and hence would also coincide with the lower baseline; Group 2 (G2), which denotes that the network is trained by means of the combination of $L_\text{pCE}$ and $L_\text{cen}$; and Group 3 (G3), for which the network is trained using the full loss function as described in (\ref{func:final_loss}).

Finally, we compare our segmentation approach with two other alternative approaches also aimed at solving the WSSS problem through a modified loss function. These loss functions are the Constrained-size Loss ($L_\text{size}$)~\cite{kervadec2019constrained} and the Seed, Expand, and Constrain (SEC) Loss ($L_\text{sec}$)~\cite{kolesnikov2016seed}: 
    \begin{align}
            L_\text{size} &= L_\text{pCE} + \lambda_\text{size}L_{\mathcal{C}(V_S)} 
            \label{func:compare_exp_loss_1}
            \\
            L_\text{sec}  &= L_\text{seed} + L_\text{expand} + L_\text{constrain}
            \label{func:compare_exp_loss_2}
    \end{align}
On the one hand, $\lambda_\text{size}$ for the $L_{\mathcal{C}(V_S)}$ term is set to $10^{-3}$. On the other hand, regarding $L_\text{sec}$, it consists of three terms, the seed loss $L_\text{seed}$, the expand loss $L_\text{expand}$, and the constrain loss $L_\text{constrain}$. In our case, we feed $L_\text{seed}$ from the scribble annotations, while, regarding $L_\text{expand}$ and $L_\text{constrain}$, we adopt the same configuration as in the original work. 

\begin{table*}[t]
    \centering
    \caption{Labels for the different experiments performed, varying the width of scribbles, the number of superpixels employed for generating the pseudo-masks, and the terms involved in the loss function employed during training. SMX stands for \textit{softmax}.}
    \label{tab:exp_define}
    \begin{tabular}{m{1.8cm}|m{3cm}|m{1.5cm}|m{1.5cm}|m{1.6cm}|m{1.5cm}|m{2.5cm}}
        \hline %\toprule
        \textbf{Configuration} & \textbf{Label} & \textbf{Scribbles width} & \textbf{Num. superpixels} & \textbf{Centroid features} & \textbf{Supervision} & \textbf{Loss function} \\
        \hline %\midrule
        \multirow{7}{1.8cm}{lower baseline}
        & E-SCR2        & 2   & -   & - & \multirow{4}{*}{only scribbles} & \multirow{4}{*}{$L_\text{pCE}$} \\
        & E-SCR5        & 5   & -   & - &  &  \\
        & E-SCR10       & 10  & -   & - &  &  \\
        & E-SCR20       & 20  & -   & - &  &  \\
        \cline{2-7} %\midrule
        & E-SCR20-SUP30 & 20  & 30  & - & \multirow{3}{*}{pseudo-masks} & \multirow{3}{*}{$L_\text{pCE}$} \\
        & E-SCR20-SUP50 & 20  & 50  & - &  &  \\
        & E-SCR20-SUP80 & 20  & 80  & - &  &  \\
        \hline %\midrule
        & E-SCR2-N      & 2     & - & SMX & \multirow{8}{*}{only scribbles} & \multirow{8}{*}{$L_\text{pCE} + L_\text{cen}\ [+ L_\text{mse}]$} \\
        & E-SCR2-NRGB   & 2     & - & SMX \& RGB   &  &  \\  
        & E-SCR5-N      & 5     & - & SMX          &  &  \\
        & E-SCR5-NRGB   & 5     & - & SMX \& RGB   &  &  \\  
        & E-SCR10-N     & 10    & - & SMX          &  &  \\
        & E-SCR10-NRGB  & 10    & - & SMX \& RGB   &  &  \\  
        & E-SCR20-N     & 20    & - & SMX          &  &  \\
        & E-SCR20-NRGB  & 20    & - & SMX \& RGB   &  &  \\  
        \cline{2-7} %\midrule
        & E-SCR20-SUP30-N       & 20  & 30 & SMX         & \multirow{6}{*}{pseudo-masks} & \multirow{6}{*}{$L_\text{pCE} + L_\text{cen}\ [+ L_\text{mse}]$}   \\
        & E-SCR20-SUP30-NRGB    & 20  & 30 & SMX \& RGB  &   &  \\  
        & E-SCR20-SUP50-N       & 20  & 50 & SMX         &   &  \\
        & E-SCR20-SUP50-NRGB    & 20  & 50 & SMX \& RGB  &   &  \\  
        & E-SCR20-SUP80-N       & 20  & 80 & SMX         &   &  \\
        & E-SCR20-SUP80-NRGB    & 20  & 80 & SMX \& RGB  &   &  \\  
        \hline %\midrule
        upper baseline
        & E-FULL                & -   & -  & -           & full mask & $L_\text{CE}$ \\
        \hline %\bottomrule
    \end{tabular}
\end{table*}

%\subsection{Discussion on the feature space} 
\subsection{About the Centroid loss feature space and the weak annotations} 
\label{sec:dis_feature}

Given the relevance that color features can have in image semantic segmentation performance~\cite{Liu2018}, the experiments reported in this section consider the incorporation of color data from the classes into the calculation and minimization of the Centroid and the MSE loss functions, $L_\text{cen}$ and $L_\text{mse}$. More specifically, we adopt a simple strategy by making use of normalized RGB features~\footnote{If $R_p = G_p = B_p = 0$, then $\text{nRGB}_p = (0,0,0)$.}:
\begin{equation}
    \text{nRGB}_p = \frac{1}{R_p+G_p+B_p}\left(R_p, G_p, B_p\right)
\end{equation}
As mentioned in Section~\ref{sec:net_arch}, the shape of $P_{cen}$ is $C\times M$, where $M = C + K$, and $K$ is a number of additional features from the classes that we incorporate into the network optimization problem. Therefore, in our case, $K = 3$. Of course, more sophisticated hand-crafted features can be incorporated into the process, though the idea of this experiment has been to make use of simple features.

Tables~\ref{tab:scribbles_compare} and~\ref{tab:superpixles_compare} evaluate the performance of our approach for different combinations of loss terms, for the two centroid feature spaces outlined before, and also depending on the kind of weak annotation that is employed as ground truth and their main feature value, i.e. width for scribbles and number of superpixels for pseudo-masks. Besides, we consider two possibilities of producing the final labelling: from the output of the segmentation network and from the clustering deriving from the predicted class centroids, i.e. label each pixel with the class label of the closest centroid; from now on, to simplify the discussion despite the language abuse, we will refer to the latter kind of output as that resulting from \textit{clustering}. Finally, Table~\ref{tab:scribbles_compare} only shows results for the visual inspection task because scribbles alone have been shown not enough for obtaining proper segmentations in the quality control case.

As can be observed in Table~\ref{tab:scribbles_compare}, segmentation and clustering mIOU for experiments E-SCR*-NRGB is lower than the mIOU for experiments E-SCR*-N, with a large gap in performance in a number of cases, what suggests that the RGB features actually do not contribute ---rather the opposite--- on improving segmentation performance when scribble annotations alone are used as supervision information for the visual inspection dataset. 

As for Table~\ref{tab:superpixles_compare}, contrary to the results shown in Table~\ref{tab:scribbles_compare}, the performance that can be observed from experiments E-SCR20-SUP*-NRGB results to be similar to that of experiments E-SCR20-SUP*-N. Additionally, the mIOU of some experiments where the integrated features, i.e. \textit{softmax} and colour, are used is even higher than if only the \textit{softmax} features are used (e.g. E-SCR20-SUP80-N/NRGB, sixth row of Table~\ref{tab:superpixles_compare}). 

At a global level, both Tables~\ref{tab:scribbles_compare} and~\ref{tab:superpixles_compare} show that our approach requires a higher number of labelled pixels to achieve higher segmentation performance when the integrated features are employed. In contrast, the use of \textit{softmax} features only requires the scribble annotations to produce good performance for the visual inspection task. Nevertheless, our approach using \textit{softmax} features achieves higher mIOU than using the integrated features in most of the experiments. As a consequence, only \textit{softmax} features are involved in the next experiments. 

\begin{sidewaystable}
    \centering
    \caption{Segmentation performance for different centroid feature spaces and different widths of the scribble annotations. \textit{*N} denotes that only the SMX (\textit{softmax}) features are used to compute $L_\text{cen}$ and $L_\text{mse}$, while \textit{*NR} denotes that the feature space for centroids prediction comprises both SMX and RGB features. \textit{Seg} denotes that the segmentation output comes directly from the segmentation network, while \textit{Clu} denotes that the segmentation output is obtained from clustering.}
    % and the *CRFs means to use dense CRFs as post-processing
    \label{tab:scribbles_compare}
%    \begin{tabular}{ p{2.cm}|p{4.5cm}||p{1.5cm}||p{1.cm}p{1.cm}p{1.cm}||p{1.5cm}|p{2.cm}||p{1.5cm}|p{2.cm}   } % ||p{1.5cm}
    \begin{tabular}{ c|c||c||ccc||c|c|c||c|c} % ||p{1.5cm}
        \toprule
        Task & Experiments & wmIOU & $L_\text{pCE}$ & $L_\text{cen}$ & $L_\text{mse}$ & mIOU (Seg) & mIOU (Seg,*N) & mIOU (Seg,*NR) & mIOU (Clu,*N) & mIOU (Clu,*NR) \\ % & *CRFs 
        \hline
        \multirow{12}{1.5cm}{Visual Inspection}
        & E-SCR2  & 0.2721 & $\checkmark$ & & & 0.3733 & - & - & - & - \\ 
        & E-SCR5  & 0.2902 & $\checkmark$ & & & 0.4621 & - & - & - & - \\ 
        & E-SCR10 & 0.3074 & $\checkmark$ & & & 0.4711 & - & - & - & - \\ 
        & E-SCR20 & 0.3233 & $\checkmark$ & & & 0.5286 & - & - & - & - \\ 
        \cline{2-11}
        % \multirow{4}{*}{Inspection}
        & E-SCR2-*  & 0.2721 & $\checkmark$ & $\checkmark$ &  & - & 0.6851 & 0.4729 & 0.6758 & 0.3889 \\   
        & E-SCR5-*  & 0.2902 & $\checkmark$ & $\checkmark$ &  & - & 0.6798 & 0.4989 & 0.6706 & 0.6020 \\   
        & E-SCR10-* & 0.3074 & $\checkmark$ & $\checkmark$ &  & - & 0.6992 & 0.5130 & 0.6710 & 0.6267 \\   
        & E-SCR20-* & 0.3233 & $\checkmark$ & $\checkmark$ &  & - & 0.6852 & 0.5562 & 0.6741 & 0.6164 \\   
        \cline{2-11}
        % \multirow{4}{*}{Inspection}
        & E-SCR2-*  & 0.2721 & $\checkmark$ & $\checkmark$ & $\checkmark$ & - & 0.6995 & 0.4724 & 0.6828 & 0.3274 \\   
        & E-SCR5-*  & 0.2902 & $\checkmark$ & $\checkmark$ & $\checkmark$ & - & 0.7134 & 0.4772 & 0.7001 & 0.2982 \\   
        & E-SCR10-* & 0.3074 & $\checkmark$ & $\checkmark$ & $\checkmark$ & - & 0.7047 & 0.4796 & 0.6817 & 0.3130 \\   
        & E-SCR20-* & 0.3233 & $\checkmark$ & $\checkmark$ & $\checkmark$ & - & 0.6904 & 0.5075 & 0.6894 & 0.6187 \\   
        \bottomrule
    \end{tabular}
\end{sidewaystable}

\begin{sidewaystable}%[htp] % [tbp]
    \centering
    \caption{Segmentation performance for different centroid feature spaces and for different amounts of superpixels to generate the pseudo-masks. \textit{*N} denotes that only the SMX (\textit{softmax}) features are used to compute $L_\text{cen}$ and $L_\text{mse}$, while \textit{*NR} denotes that the feature space comprises both SMX and RGB features. \textit{Seg} denotes that the segmentation output comes directly from the segmentation network, while \textit{Clu} denotes that the segmentation output is obtained from clustering.}
    % and the *CRFs means to use dense CRFs as post-processing
    \label{tab:superpixles_compare}
%    \begin{tabular}{ p{2.cm}|p{4.5cm}||p{1.5cm}||p{1.cm}p{1.cm}p{1.cm}||p{1.5cm}|p{2.cm}||p{1.5cm}|p{2.cm}   } % ||p{1.5cm}
    \begin{tabular}{ c|c||c||ccc||c|c|c||c|c } % ||p{1.5cm}
        \toprule
        % Task & Experiments & wmIOU & $L_\text{pCE}$ & $L_\text{cen}$ & $L_\text{mse}$ & mIOU (Seg,*C) & mIOU (Seg,*R,*C) & mIOU (Clu,*C) & mIOU (Clu,*R,*C) \\ % & *CRFs 
        Task & Experiments & wmIOU & $L_\text{pCE}$ & $L_\text{cen}$ & $L_\text{mse}$ & mIOU (Seg) & mIOU (Seg,*N) & mIOU (Seg,*NR) & mIOU (Clu,*N) & mIOU (Clu,*NR) \\ % & *CRFs 
        \hline
        \multirow{9}{1.5cm}{Visual Inspection}
        & E-SCR20-SUP30 & 0.6272 & $\checkmark$ & & & 0.6613 & - & - & - & -\\ 
        & E-SCR20-SUP50 & 0.6431 & $\checkmark$ & & & 0.7133 & - & - & - & -\\ 
        & E-SCR20-SUP80 & 0.6311 & $\checkmark$ & & & 0.7017 & - & - & - & -\\ 
        \cline{2-11}
        & E-SCR20-SUP30-* & 0.6272 & $\checkmark$ & $\checkmark$ &  & - & 0.6848 & 0.6847 & 0.7081 & 0.6859\\   
        & E-SCR20-SUP50-* & 0.6431 & $\checkmark$ & $\checkmark$ &  & - & 0.7447 & 0.7368 & 0.7372 & 0.7136\\   
        & E-SCR20-SUP80-* & 0.6311 & $\checkmark$ & $\checkmark$ &  & - & 0.7242 & 0.7355 & 0.7127 & 0.6761\\   
        \cline{2-11}
        & E-SCR20-SUP30-* & 0.6272 & $\checkmark$ & $\checkmark$ & $\checkmark$ & - & 0.6919 & 0.7071 & 0.6987 & 0.7076 \\   
        & E-SCR20-SUP50-* & 0.6431 & $\checkmark$ & $\checkmark$ & $\checkmark$ & - & 0.7542 & 0.7133 & 0.7491 & 0.7294 \\   
        & E-SCR20-SUP80-* & 0.6311 & $\checkmark$ & $\checkmark$ & $\checkmark$ & - & 0.7294 & 0.7246 & 0.7268 & 0.7118 \\   
        % \hline 
        \midrule[.6pt]
        \multirow{9}{1.5cm}{\raggedright Quality Control}
        & E-SCR20-SUP30 & 0.4710 & $\checkmark$ & & & 0.5419 & - & - & - & -\\ 
        & E-SCR20-SUP50 & 0.5133 & $\checkmark$ & & & 0.6483 & - & - & - & -\\ 
        & E-SCR20-SUP80 & 0.5888 & $\checkmark$ & & & 0.7015 & - & - & - & -\\ 
        \cline{2-11}
        & E-SCR20-SUP30-* & 0.4710 & $\checkmark$ & $\checkmark$ &  & - & 0.6882 & 0.6889 & 0.6142 & 0.6062\\   
        & E-SCR20-SUP50-* & 0.5133 & $\checkmark$ & $\checkmark$ &  & - & 0.7236 & 0.7203 & 0.6644 & 0.6480\\   
        & E-SCR20-SUP80-* & 0.5888 & $\checkmark$ & $\checkmark$ &  & - & 0.7594 & 0.7337 & 0.6768 & 0.6451\\   
        \cline{2-11}
        & E-SCR20-SUP30-* & 0.4710 & $\checkmark$ & $\checkmark$ & $\checkmark$ & - & 0.7030 & 0.6237 & 0.5910 & 0.6077 \\   
        & E-SCR20-SUP50-* & 0.5133 & $\checkmark$ & $\checkmark$ & $\checkmark$ & - & 0.7291 & 0.7046 & 0.6605 & 0.6372 \\   
        & E-SCR20-SUP80-* & 0.5888 & $\checkmark$ & $\checkmark$ & $\checkmark$ & - & 0.7679 & 0.7409 & 0.6687 & 0.6780 \\   
        \bottomrule
    \end{tabular}
\end{sidewaystable}

\subsection{Effect of the loss function terms}
\label{sec:effect_cen}

This section considers the effect of $L_\text{cen}$ and $L_\text{mse}$ on the segmentation results by analyzing performance of experiments in groups G1, G2 and G3. From Table~\ref{tab:scribbles_compare}, one can see that the mIOU of experiments in G2 is significantly higher than that of experiments in G1, where the maximum gap in mIOU between G1 and G2 is 0.3118 (E-SCR2 and E-SCR2-N). As for the segmentation performance for G3 experiments, it is systematically above that of G2 experiments for the same width of the scribble annotations and if centroids are built only from the \textit{softmax} features. When the colour data is incorporated, segmentation performance decreases from G2 to G3.

Table~\ref{tab:superpixles_compare} also shows performance improvements from G2 experiments, i.e. when $L_\text{cen}$ is incorporated into the loss function, over the performance observed from experiments in G1, and, in turn, segmentation results from G3 experiments are superior to that of G2 experiments, and this is observed for both tasks. Therefore, the incorporation of the $L_\text{cen}$ and $L_\text{mse}$ terms into the loss function benefits performance, gradually increasing the mIOU of the resulting segmentations.

Regarding the segmentation computed from clustering, the mIOU of experiments in G3 is also higher than that of experiments in G2. In addition, it can be found out in Table~\ref{tab:scribbles_compare} and Table~\ref{tab:superpixles_compare} that the mIOU from clustering in some G2 experiments is slightly higher than that for G3 experiments  (E-SCR20-SUP30-N on both tasks and specifically E-SCR20-SUP80-N for the quality control task), while the mIOU from segmentation in G2 is lower than that of G3. In other words, it seems that $L_\text{mse}$, in some cases, makes the segmentation quality from clustering deteriorate.

Overall, the incorporation of $L_\text{cen}$ and $L_\text{mse}$ improves segmentation performance for both tasks, and labelling from segmentation turns out to be superior to that deriving from class centroids. 

\subsection{Impact of weak annotations and their propagation}
\label{sec:influence_weak}

\begin{sidewaysfigure*} % [htb]
    % \begin{adjustwidth}{-1.5cm}{}
    \centering
    % \begin{tabular}{@{\hspace{-10mm}}c@{\hspace{1mm}}c@{\hspace{1mm}}c@{\hspace{0mm}}}
    \begin{tabular}{@{\hspace{0mm}}c@{\hspace{0mm}}c@{\hspace{0mm}}c@{\hspace{0mm}}}
        \includegraphics[width=0.34\textwidth,height=0.30\textwidth]{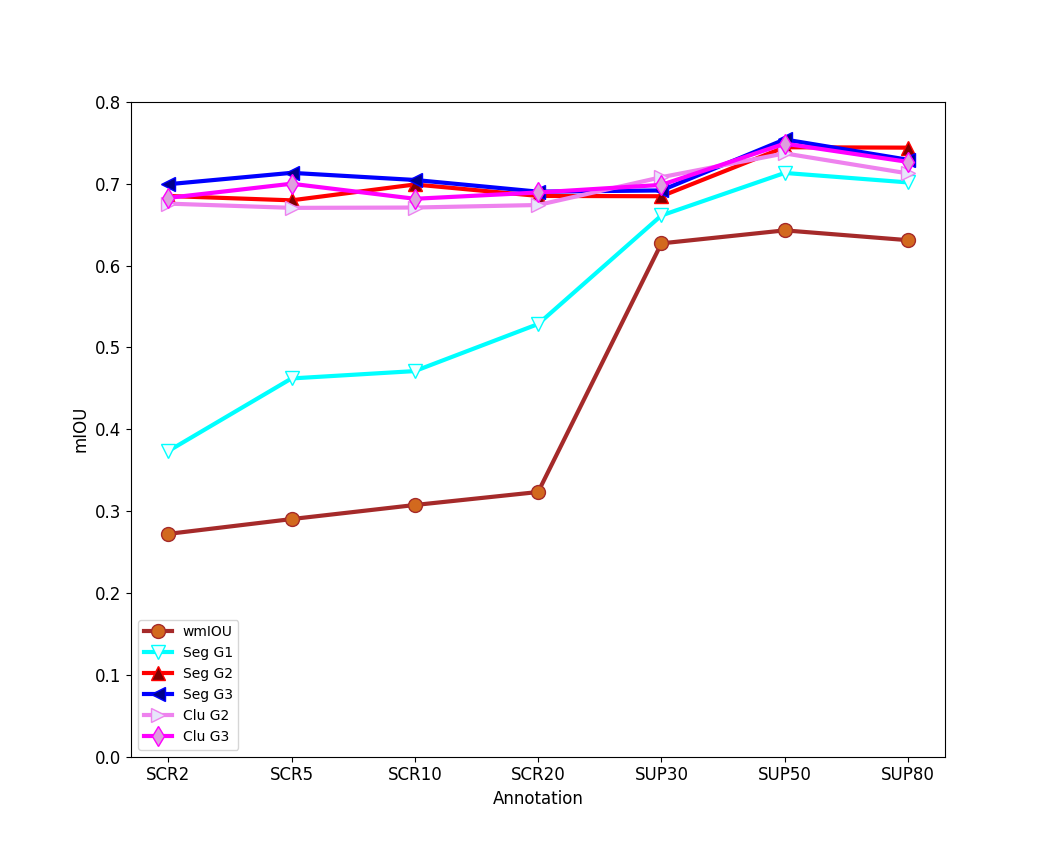}
        &
        \includegraphics[width=0.34\textwidth,height=0.30\textwidth]{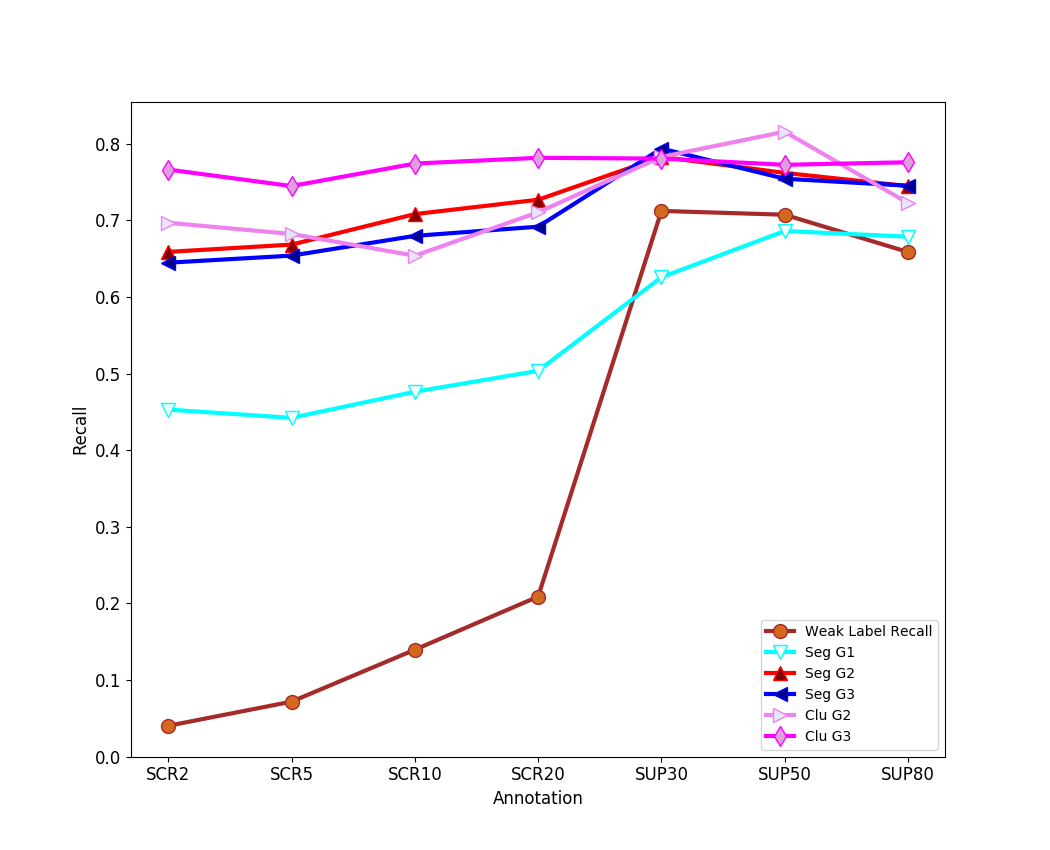}
        &
        \includegraphics[width=0.34\textwidth,height=0.30\textwidth]{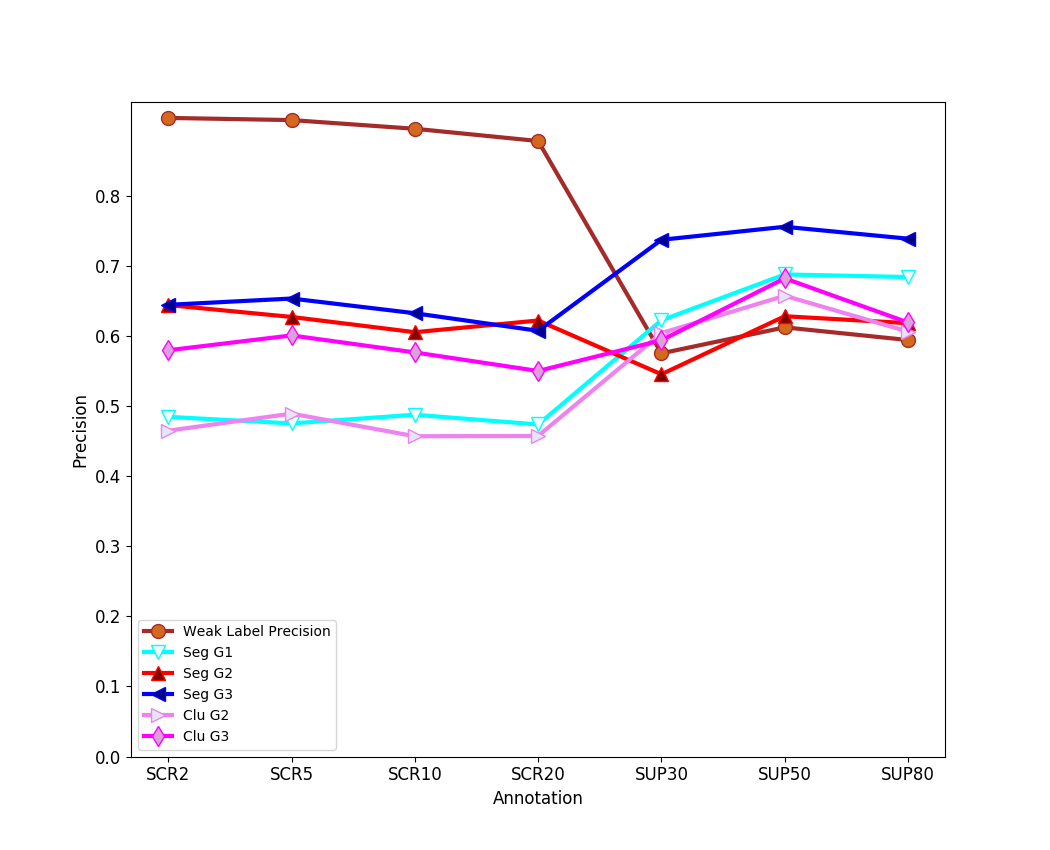} \\
        \footnotesize (a) & \footnotesize (b) & \footnotesize (c) \\
        \includegraphics[width=0.34\textwidth,height=0.30\textwidth]{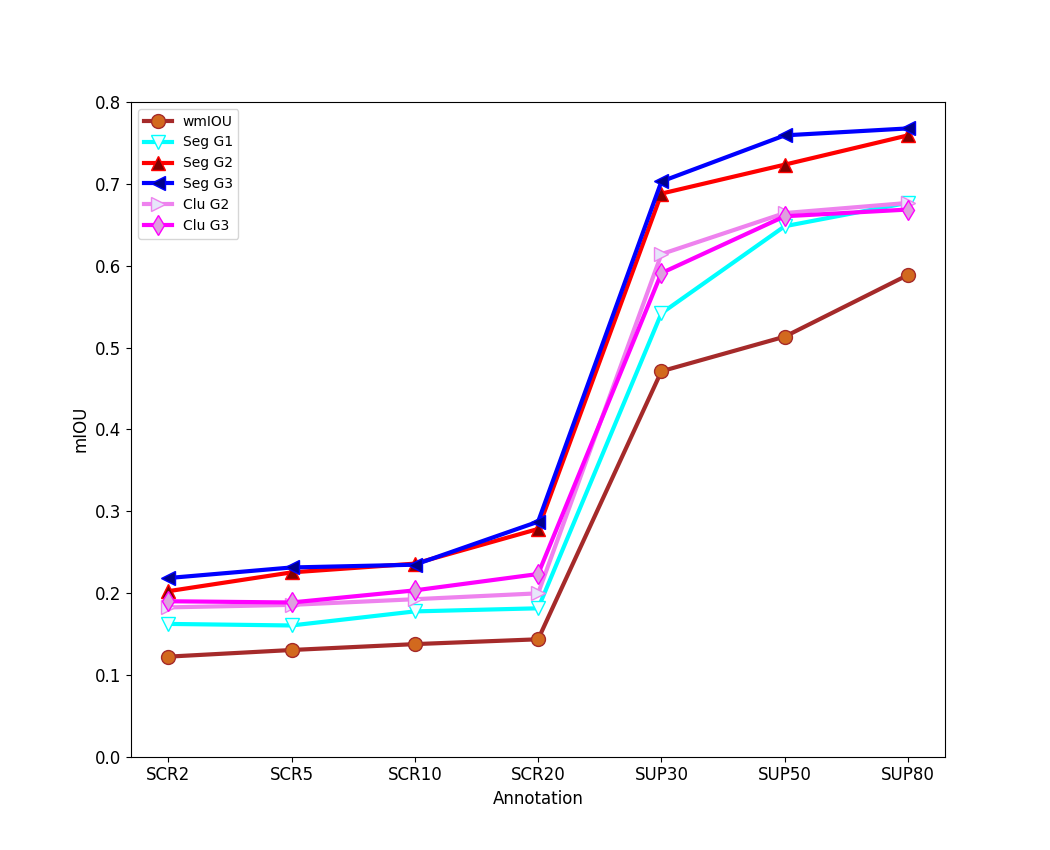}
        &
        \includegraphics[width=0.34\textwidth,height=0.30\textwidth]{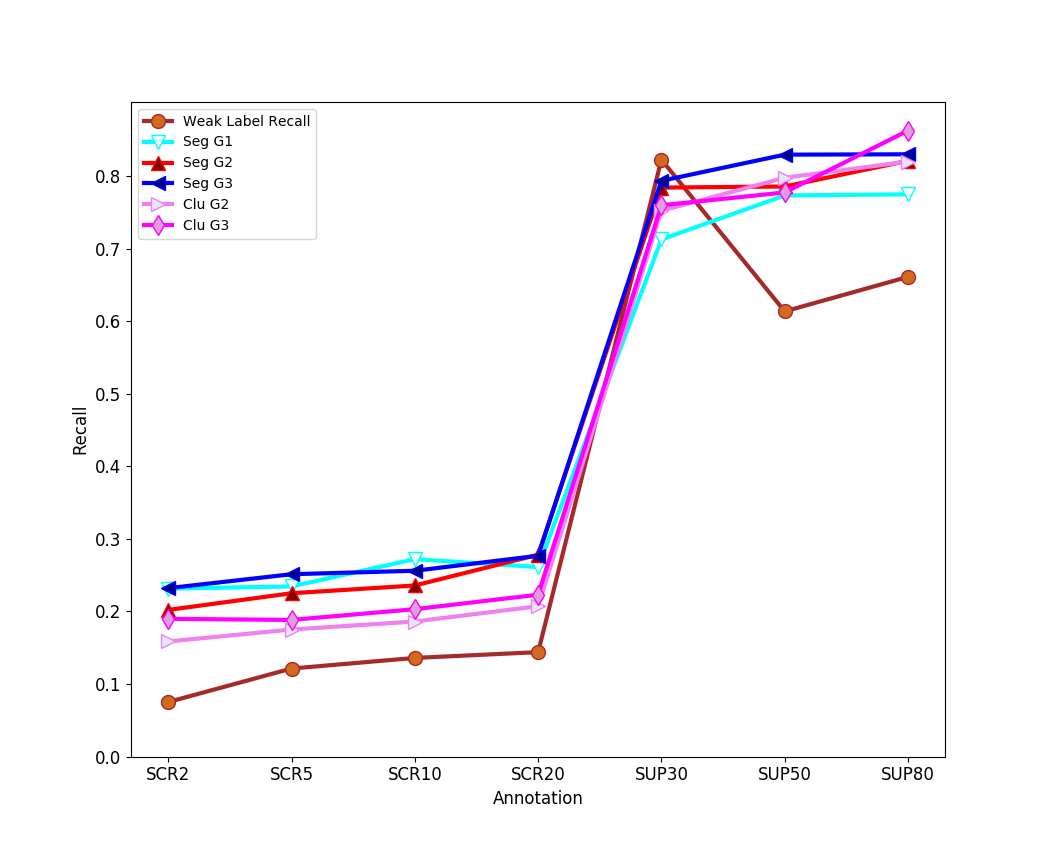}
        &
        \includegraphics[width=0.34\textwidth,height=0.30\textwidth]{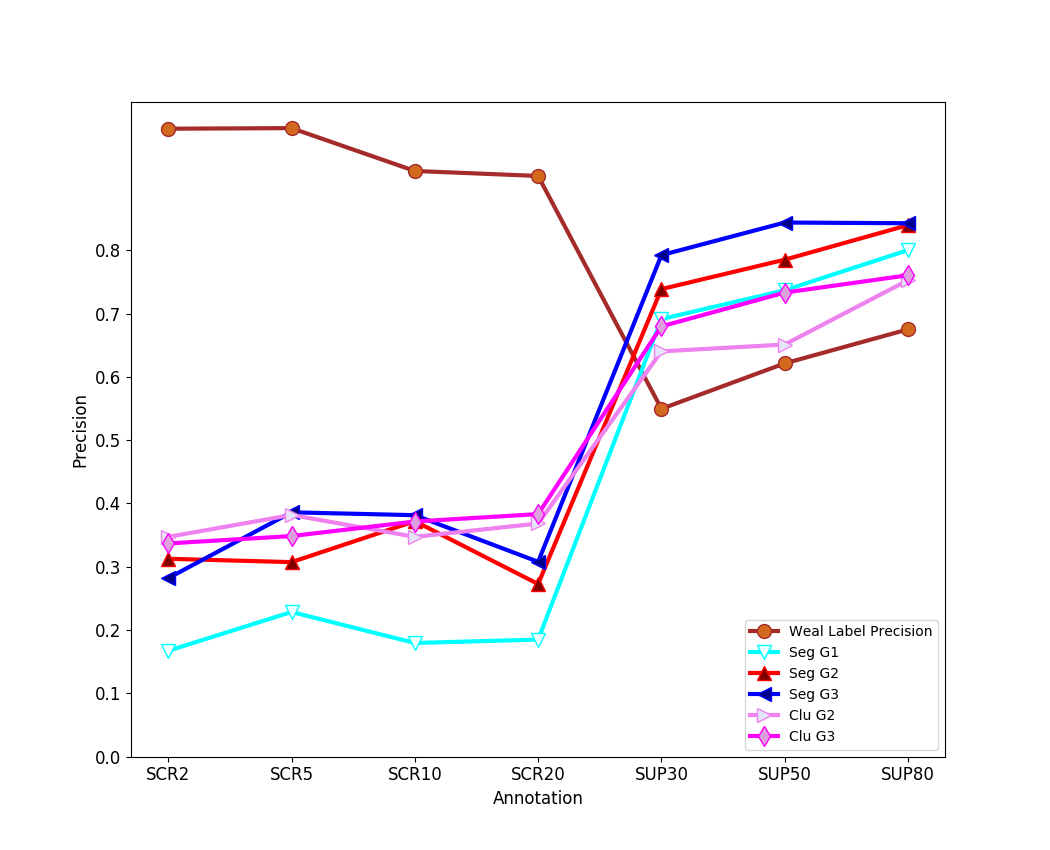} \\
        \footnotesize (d) & \footnotesize (e) & \footnotesize (f)
    \end{tabular}
    \caption{Performance metrics for our approach under different sorts of weak annotations. The first row plots are for the visual inspection task, while those of the second row are for the quality control task. In both rows, from left to right, the three figures plot respectively the mIOU, the mean Recall, and the mean Precision. SUP30, SUP50 and SUP80 labels correspond to the use of 20 pixel-wide scribbles.}
    \label{fig:influence_weak_annotation}
    % \end{figure*}
\end{sidewaysfigure*}

In this section, we evaluate our approach under different weak annotations and their propagation, and discuss on their impact on segmentation performance for both tasks. To this end, we plot in Fig.~\ref{fig:influence_weak_annotation} the mIOU (complementarily to Tables~\ref{tab:scribbles_compare} and~\ref{tab:superpixles_compare}), recall and precision values resulting after the supervision of different sorts of weak annotations for the two tasks. A first analysis of these plots reveals that the curves corresponding to the G3 experiments are above than those for G1 and G2 groups for all the performance metrics considered.

Regarding the visual inspection task, Fig.~\ref{fig:influence_weak_annotation}(a) shows that the mIOU values for the G2 and G3 groups are above those for G1 (the lower baseline), which follows a similar shape as the wmIOU values, while those from G2 and G3 groups keep at a more or less constant level for the different sorts of weak annotations. As for the quality control task, the mIOU values for all groups are similar among all groups and similar to the wmIOU values, as shown in Fig. \ref{fig:influence_weak_annotation}(d). Globally, this behaviour clearly shows that the scribbles are enough for describing the classes in the case of the visual inspection task, which is a binary classification problem, while this is not true for the quality control class, a multi-class problem, and this makes necessary resort to the pseudo-masks (G2 and G3 groups) to achieve a higher performance. The fact that for both tasks the lower baseline (G1 group) always achieves lower mIOU values also corroborates the relevance of the Centroid Loss, despite its ultimate contribution to the segmentation performance is also affected by the quality of the weak annotations involved, i.e. the pseudo-masks deriving from scribbles and superpixels for the cases of the G2 and G3 groups.

Additionally, observing the precision curves shown in Fig.~\ref{fig:influence_weak_annotation}(c) and (f), one can notice that the precision for exclusively the weak annotations show a sharp decline when the weak annotations shift from scribbles to pseudo-masks. As can be noticed from the pseudo-masks shown in the second and third rows of Fig.~\ref{fig:pseudo_mask_exp}, when the number of superpixels is low, e.g. 30, the pseudo-masks contain an important number of incorrectly labelled pixels, significantly more than that of the scribble annotations, and this is the reason for the aforementioned decline. The recall curves, however, exhibit an upward trend as can be observed in Fig.~\ref{fig:influence_weak_annotation}(b) and (e) because of the larger amount of information ultimately provided by the weak annotations. On the other side, we can also notice that, in general, precision and recall values are higher for the G3 group than for the G2 group, and both curves are above those for the G1 group, and this behaviour replicates for the two tasks. Finally, the output from clustering does not clearly lead to a different performance, better or worse, over the alternative outcome from the segmentation network, showing that clustering is less appropriate for the quality control task from the point of view of the recall metric.

From a global perspective, all this suggests that (a) segmentation quality benefits from the use of pseudo-masks, (b) overcoming always the lower baseline based on the use of exclusively scribbles, (c) despite the incorrectly labelled pixels contained in pseudo-masks, (d) provided that the proper loss function is adopted, e.g. the full loss expressed in (\ref{func:final_loss}), which in particular (e) comprises the Centroid Loss.

%\subsection{Comparative results}
\subsection{Comparison with other loss functions}
\label{sec:performance_compare}

In Table~\ref{tab:previous_compare}, we compare the segmentation performance of our approach for the two tasks with that resulting from the use of the Constrained-size Loss $L_\text{size}$~\cite{kervadec2019constrained} and the SEC Loss $L_\text{sec}$~\cite{kolesnikov2016seed} for different variations of weak annotations. As for the visual inspection task, the network trained with $L_\text{sec}$ is clearly inferior to the one resulting for our loss function, and the same can be said for $L_\text{size}$, although, in this case, the performance gap is shorter, even negligible when the width of the scribbles is of 20 pixels. When the pseudo-masks are involved, our approach is also better though the difference with both $L_\text{size}$ and $L_\text{sec}$ is more reduced. Regarding the quality control task, Table~\ref{tab:previous_compare} shows that our approach overcomes both, at a significant distance, far more than for the visual inspection task. 

Summing up, we can conclude that the loss function proposed in (\ref{func:final_loss}) outperforms both the Constrained-size Loss $L_\text{size}$ and the SEC Loss $L_\text{sec}$ on the visual inspection and the quality control tasks.

\begin{table}
    \centering
    % \caption{Performance comparison on our two tasks using different approaches. Best performance are indicated in bold.}
    \caption{Comparison of different loss functions for both the visual inspection and the quality control tasks. mIOU values are provided. Best performance is highlighted in bold.}
    \label{tab:previous_compare}
    % \begin{tabular}{ p{1.6cm}|p{3cm}|p{2cm}|p{3.3cm}|p{1.5cm} } % |p{3cm}
    \begin{tabular}{@{\hspace{1mm}}l|c|c|c|c@{\hspace{1mm}}} % |p{3cm}
        \toprule
        Task & Weak Annotation & $L_\text{size}$~\cite{kervadec2019constrained} & $L_\text{sec}$~\cite{kolesnikov2016seed} & Ours \\ % & Weak Annotation 
        \hline
        % \multirow{9}{1.5cm}{Inspection}
        \multirow{9}{1.1cm}[3.2mm]{Visual Inspection}
        & E-SCR2-N      & 0.6098          & 0.4366          & \textbf{0.6995}          \\  % & E-SCR2-N        
        & E-SCR5-N      & 0.6537          & 0.4372          & \textbf{0.7134}          \\  % & E-SCR5-N        
        & E-SCR10-N     & 0.6754          & 0.5486          & \textbf{0.7047}          \\  % & E-SCR10-N       
        & E-SCR20-N     & \textbf{0.6909} & 0.5624          & 0.6904                   \\  % & E-SCR20-N       
        \cline{2-5}
        & E-SCR20-SUP30-N & \textbf{0.7068} & 0.6397          & 0.6919                   \\  % & E-SCR20-SUP30-N 
        & E-SCR20-SUP50-N & 0.6769          & 0.7428          & \textbf{0.7542}          \\  % & E-SCR20-SUP50-N 
        & E-SCR20-SUP80-N & 0.7107          & 0.6546          & \textbf{0.7294}          \\  % & E-SCR20-SUP80-N 
        % \hline
        \midrule[.6pt]
        % \multirow{3}{1.5cm}{Quality Control}
        \multirow{3}{1.1cm}[0mm]{Quality Control}
        & E-SCR20-SUP30-N & 0.4724          & 0.5808          & \textbf{0.7030}          \\  % & E-SCR20-SUP30-N 
        & E-SCR20-SUP50-N & 0.4985          & 0.6262          & \textbf{0.7291}          \\  % & E-SCR20-SUP50-N 
        & E-SCR20-SUP80-N & 0.5051          & 0.6918          & \textbf{0.7679}          \\  % & E-SCR20-SUP80-N 
        \bottomrule
    \end{tabular}
\end{table}
%\subsection{Experimental Results}
\subsection{Final tuning and results}
\label{sec:result_displays}

As has been already highlighted along the previous sections, the network trained by means of the loss function described in (\ref{func:final_loss}), which in particular comprises the Centroid Loss, attains the best segmentation performance against other approaches and for the two tasks considered in this work. In order to check whether segmentation performance can increase further, in this section we incorporate a dense CRF as a post-processing stage of the outcome of the network. Table~\ref{tab:annotation_compare} collects metric values for the final performance attained by the proposed WSSS method and as well by the upper baseline method (E-FULL). To assess the influence of the CRF-based stage, in Table~\ref{tab:annotation_compare}, we report mIOU, precision and recall values, together with the F$_1$ score, as well as the wmIOU values.

Regarding the visual inspection task, Table~\ref{tab:annotation_compare} shows that case E-SCR20-SUP50-N leads to the best segmentation mIOU (0.7542). After dense CRF, the mIOU reaches a value of 0.7859, with a performance gap with E-FULL of 0.0474. Case E-SCR20-SUP30-N attains the highest recall (0.7937), but the corresponding precision (0.7081) and F$_1$ score (0.7485) are not the highest; the mIOU is also the second lowest (0.6919). This is because the segmentation result for E-SCR20-SUP30-N contains more incorrect predictions than E-SCR20-SUP50-N. Consequently, a configuration of 20-pixel scribbles and 50 superpixels for pseudo-mask generation leads to the best performance, with a slightly increase thanks to the CRF post-processing stage. The outcome from clustering is not far in quality to those values, but, as can be observed, it is not as good (the best mIOU and F$_1$ scores are, respectively, 0.7491 and 0.7250).

As for the quality control task, the E-SCR20-SUP80-N case reaches the highest mIOU (0.7679) among all cases, with the second best F$_1$ (0.8350). For this task, the precision metric highlights a different case, E-SCR20-SUP50-N, as the best configuration which as well attains the largest F$_1$ score, though at a very short distance to the E-SCR20-SUP80-N case. After dense CRF, the final segmentation mIOU is 0.7707. The most adequate configuration seems to be, hence, 20-pixel scribbles and 80 superpixels for psedudo-mask generation. The gap in this case with regard to full supervision is 0.0897. Similarly to the visual inspection task, results from clustering are close in accuracy to the previous levels, but not better (for this task, the best mIOU and F$_1$ scores are, respectively, 0.6687 and 0.8086).

From a global perspective, the results obtained indicate that 20-pixel scribbles, together with a rather higher number of superpixels, so that they adhere better to object boundaries, are the best options for both tasks. In comparison with the lower baseline (G1 group), the use of the full loss function, involving the Centroid Loss, clearly makes training improve segmentation performance significantly, with a slight decrease regarding full supervision. Segmentation results deriving from clustering are not better for any of the tasks considered. 

\begin{sidewaysfigure*}[htb]
    \centering
    % org image & gt & fully & scribbles & pseudo 
    % \begin{tabular}{@{\hspace{-20mm}}c@{\hspace{1mm}}c@{\hspace{1mm}}c@{\hspace{1mm}}c@{\hspace{1mm}}c@{\hspace{1mm}}c@{\hspace{1mm}}c@{\hspace{1mm}}}
    \begin{tabular}{@{\hspace{0mm}}c@{\hspace{1mm}}c@{\hspace{1mm}}c@{\hspace{1mm}}c@{\hspace{1mm}}c@{\hspace{1mm}}c@{\hspace{1mm}}c@{\hspace{0mm}}}
        \includegraphics[width=30mm,height=30mm]{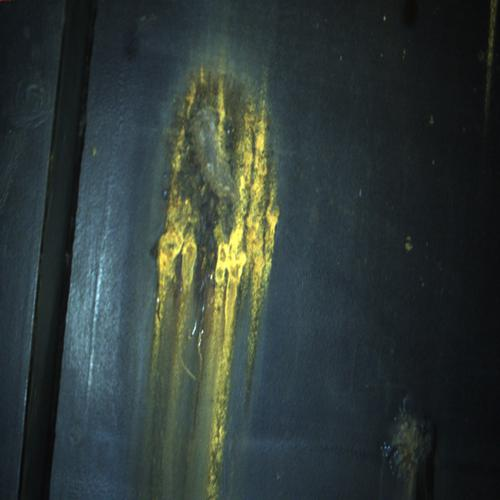}
        &
        \includegraphics[width=30mm,height=30mm]{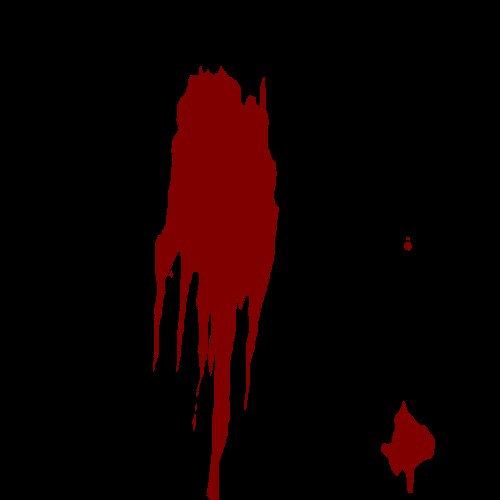}
        &
        \includegraphics[width=30mm,height=30mm]{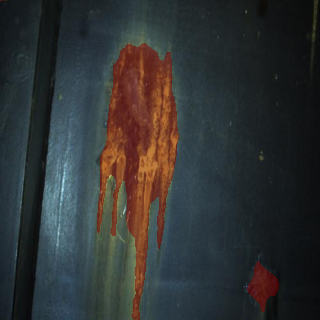}
        &
        \includegraphics[width=30mm,height=30mm]{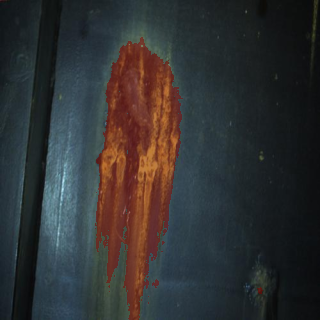}
        &
        \includegraphics[width=30mm,height=30mm]{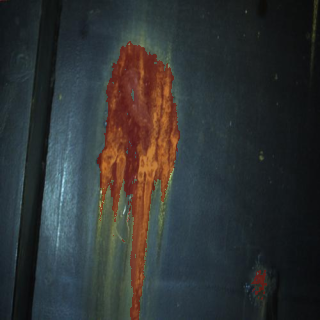}
        &
        \includegraphics[width=30mm,height=30mm]{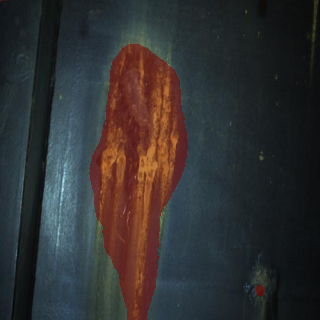}
        &
        \includegraphics[width=30mm,height=30mm]{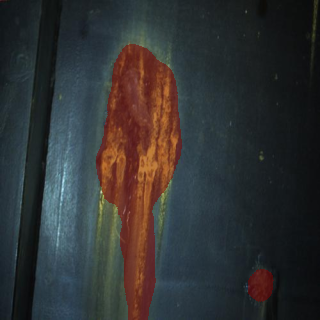}
        \\
        
        \includegraphics[width=30mm,height=30mm]{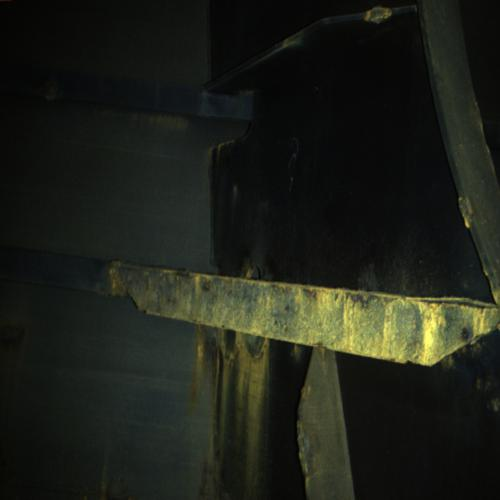}
        &
        \includegraphics[width=30mm,height=30mm]{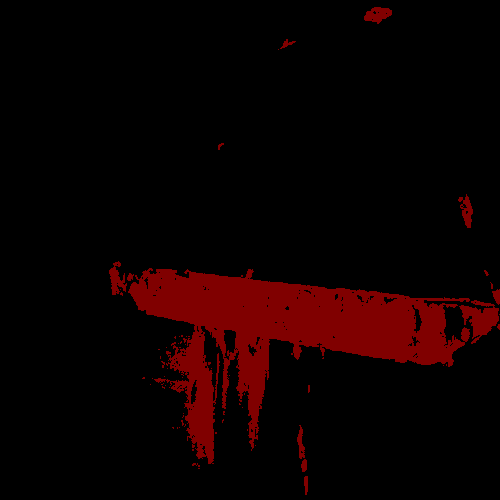}
        &
        \includegraphics[width=30mm,height=30mm]{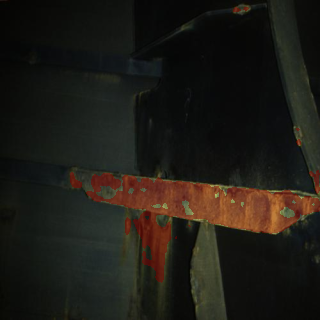}
        &
        \includegraphics[width=30mm,height=30mm]{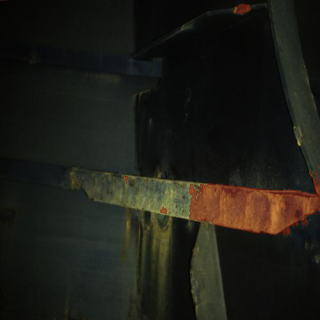}
        &
        \includegraphics[width=30mm,height=30mm]{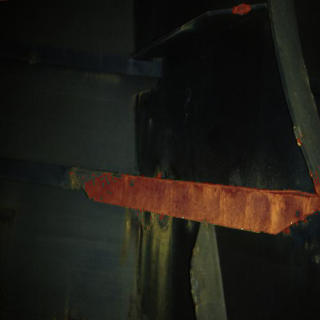}
        &
        \includegraphics[width=30mm,height=30mm]{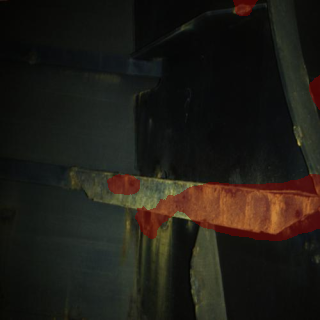}
        &
        \includegraphics[width=30mm,height=30mm]{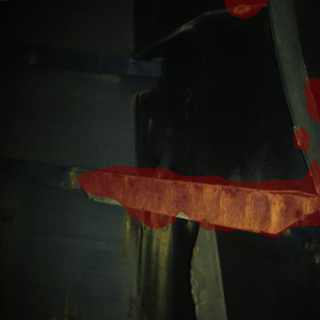}
        
        \\
        \includegraphics[width=30mm,height=30mm]{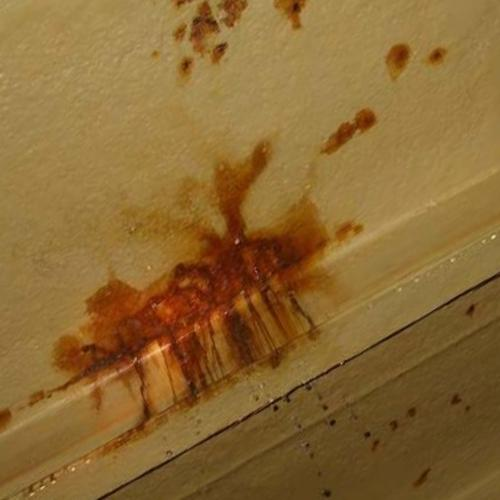}
        &
        \includegraphics[width=30mm,height=30mm]{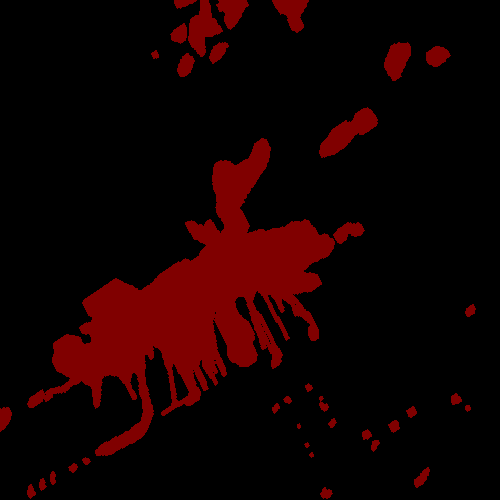}
        &
        \includegraphics[width=30mm,height=30mm]{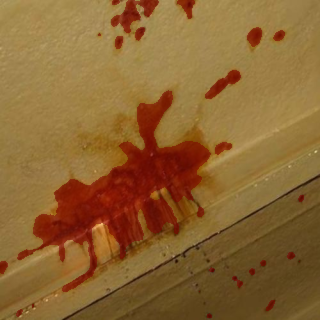}
        &
        \includegraphics[width=30mm,height=30mm]{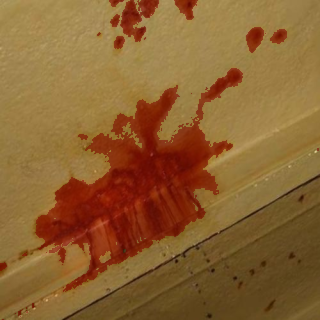}
        &
        \includegraphics[width=30mm,height=30mm]{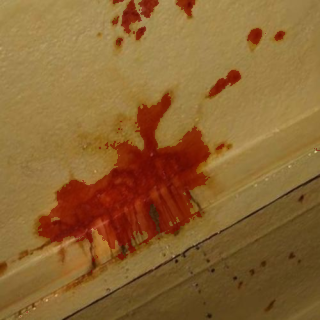}
        &
        \includegraphics[width=30mm,height=30mm]{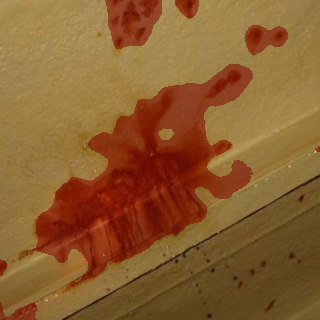}
        &
        \includegraphics[width=30mm,height=30mm]{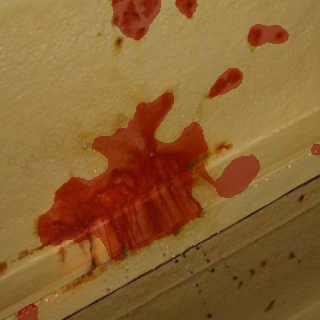}
        \\
        
        \includegraphics[width=30mm,height=30mm]{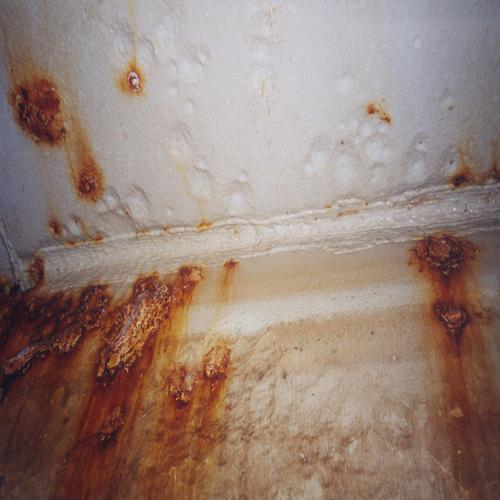}
        &
        \includegraphics[width=30mm,height=30mm]{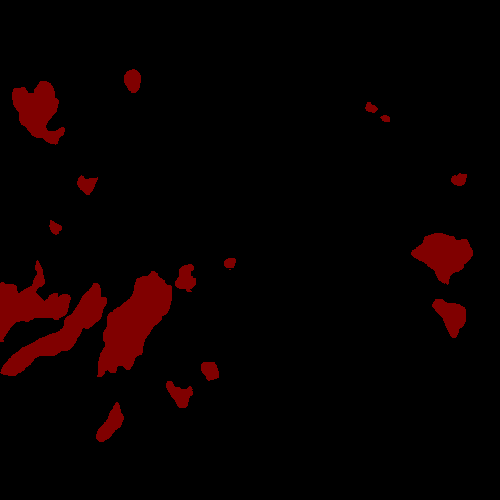}
        &
        \includegraphics[width=30mm,height=30mm]{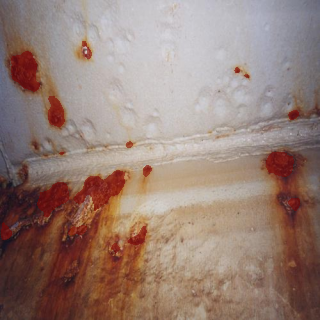}
        &
        \includegraphics[width=30mm,height=30mm]{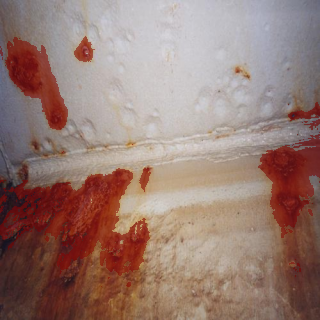}
        &
        \includegraphics[width=30mm,height=30mm]{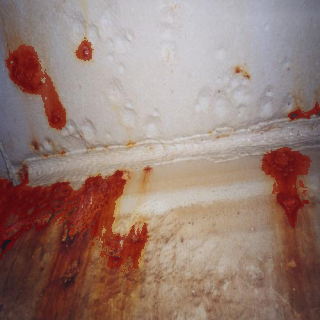}
        &
        \includegraphics[width=30mm,height=30mm]{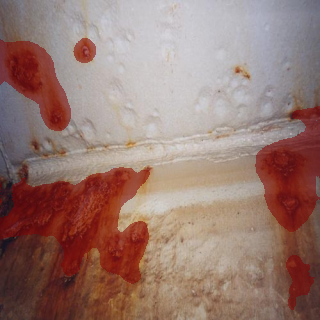}
        &
        \includegraphics[width=30mm,height=30mm]{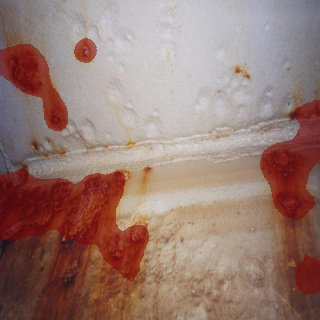}
        \\
        % Image & Ground Truth & Full Supervision & E-SCR20-N (seg) & E-SCR20-SUP50-N (seg) & E-SCR20-N (clu) & E-SCR20-SUP50-N (clu)
        \footnotesize original image & 
        \footnotesize full mask & 
        \footnotesize E-FULL & 
        \footnotesize E-SCR20-N (seg) & 
        \footnotesize E-SCR20-SUP50-N (seg) & 
        \footnotesize E-SCR20-N (clu) & 
        \footnotesize E-SCR20-SUP50-N (clu)
    \end{tabular}
    \caption{Examples of segmentation results for the visual inspection task: (1st column) original images, (2nd column) full mask, (3rd column) results of the fully supervised approach, (4th \& 5th columns) segmentation output for E-SCR20-N and E-SCR20-SUP50-N after dense CRF, (6th \& 7th columns) segmentation output from clustering for the same configurations.}
    \label{fig:seg_results}
    % \end{figure*}
\end{sidewaysfigure*}

\begin{figure*}[htb]
    \centering
    % org image & gt & fully & scribbles & pseudo 
    \begin{tabular}{@{\hspace{0mm}}c@{\hspace{1mm}}c@{\hspace{1mm}}c@{\hspace{1mm}}c@{\hspace{1mm}}c@{\hspace{1mm}}}
        \includegraphics[width=30mm,height=30mm]{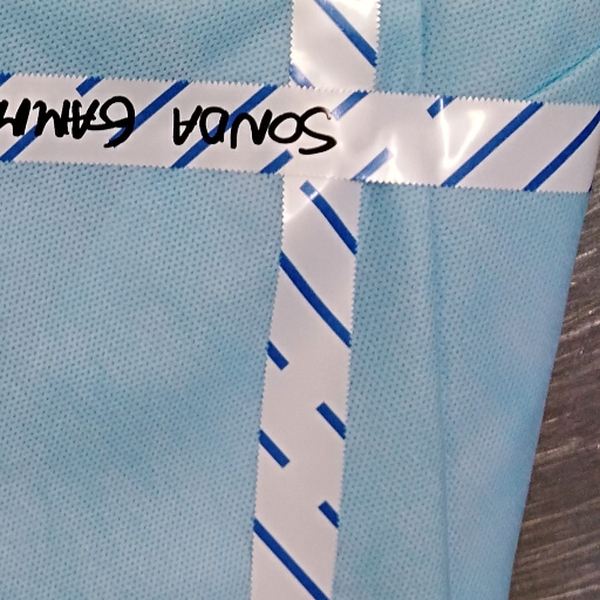}
        &
        \includegraphics[width=30mm,height=30mm]{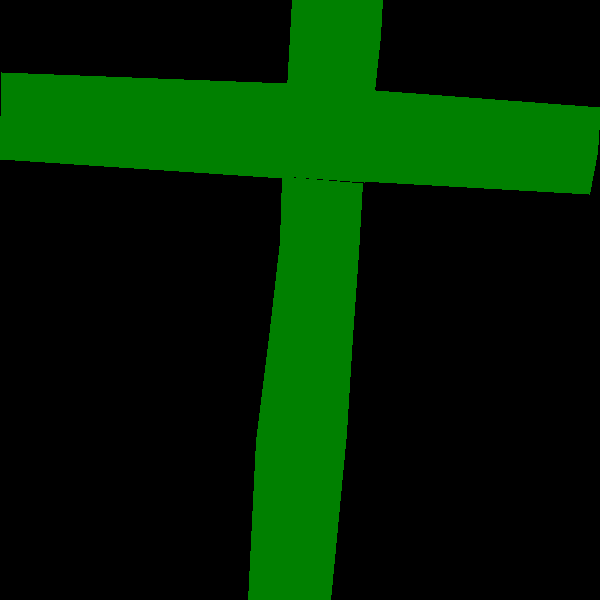}
        &
        \includegraphics[width=30mm,height=30mm]{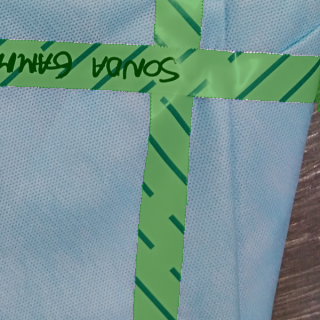}
        &
        \includegraphics[width=30mm,height=30mm]{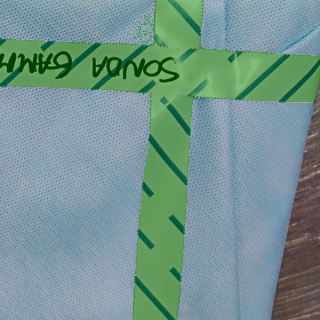}
        &
        \includegraphics[width=30mm,height=30mm]{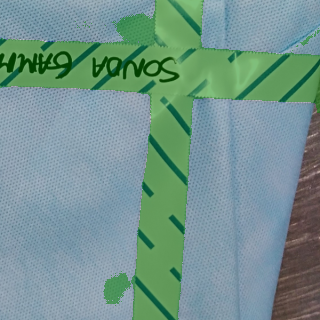}
        \\
        
        \includegraphics[width=30mm,height=30mm]{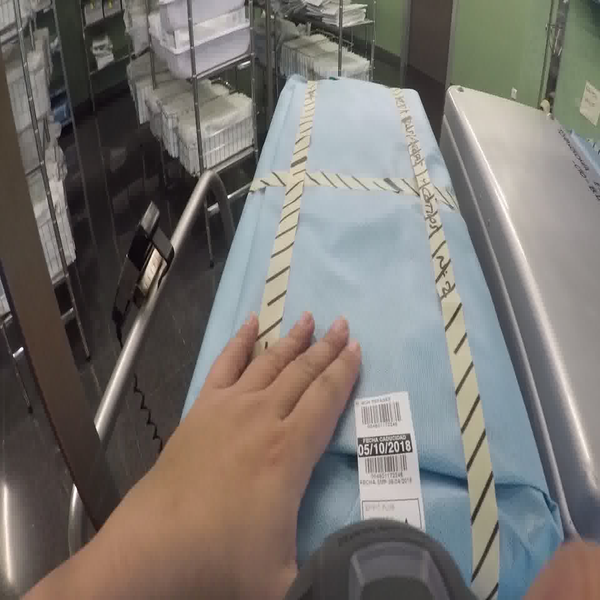}
        &
        \includegraphics[width=30mm,height=30mm]{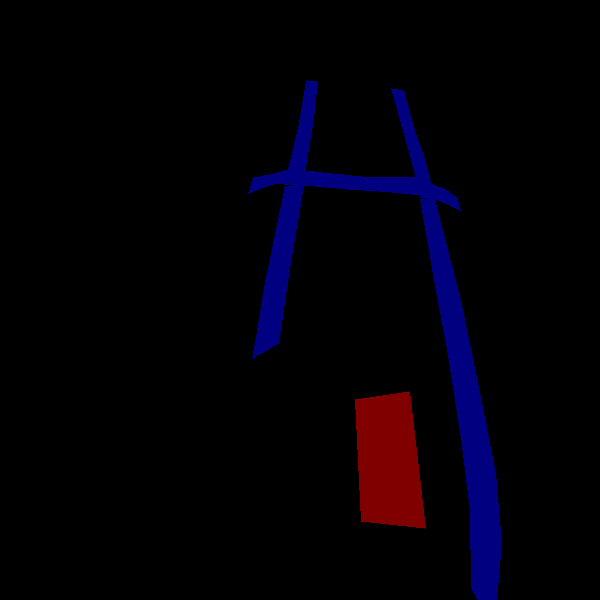}
        &
        \includegraphics[width=30mm,height=30mm]{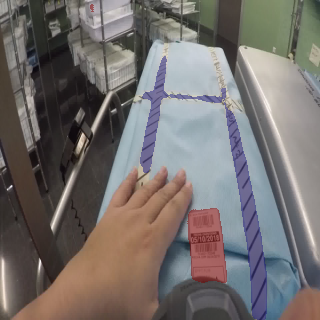}
        &
        \includegraphics[width=30mm,height=30mm]{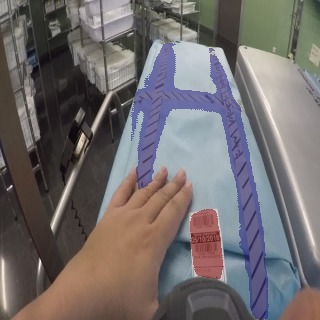}
        &
        \includegraphics[width=30mm,height=30mm]{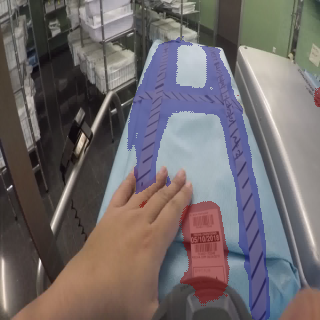}
        \\
        
        \includegraphics[width=30mm,height=30mm]{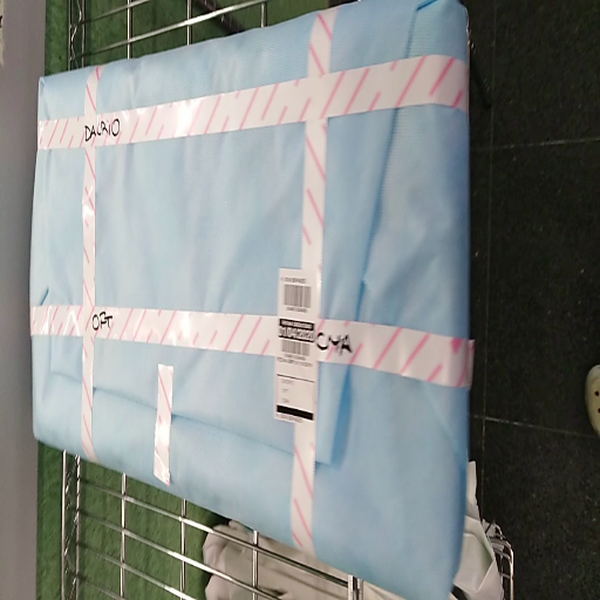}
        &
        \includegraphics[width=30mm,height=30mm]{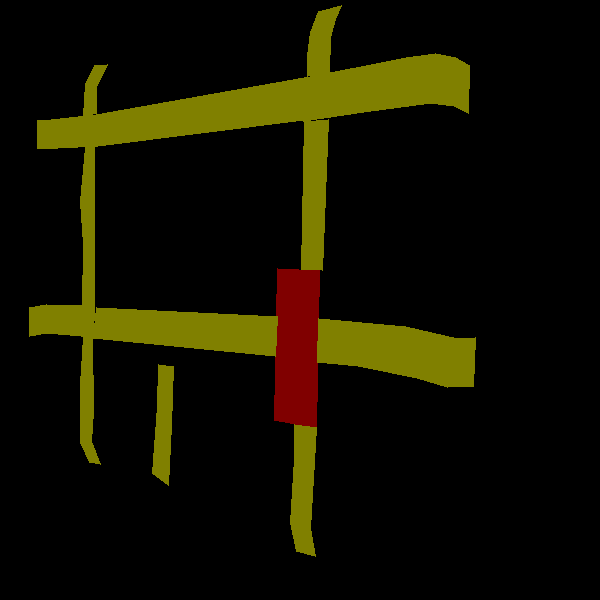}
        &
        \includegraphics[width=30mm,height=30mm]{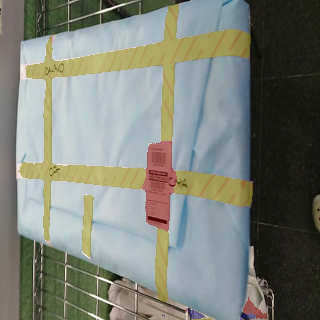}
        &
        \includegraphics[width=30mm,height=30mm]{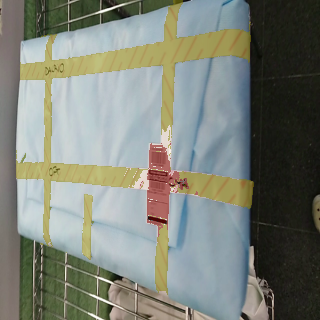}
        &
        \includegraphics[width=30mm,height=30mm]{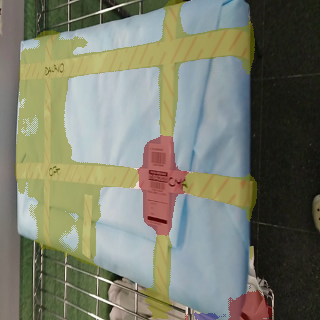}
        \\
        
        \includegraphics[width=30mm,height=30mm]{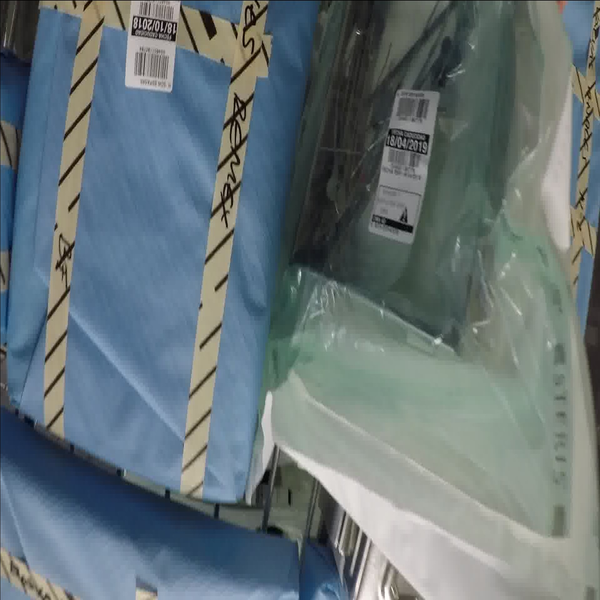}
        &
        \includegraphics[width=30mm,height=30mm]{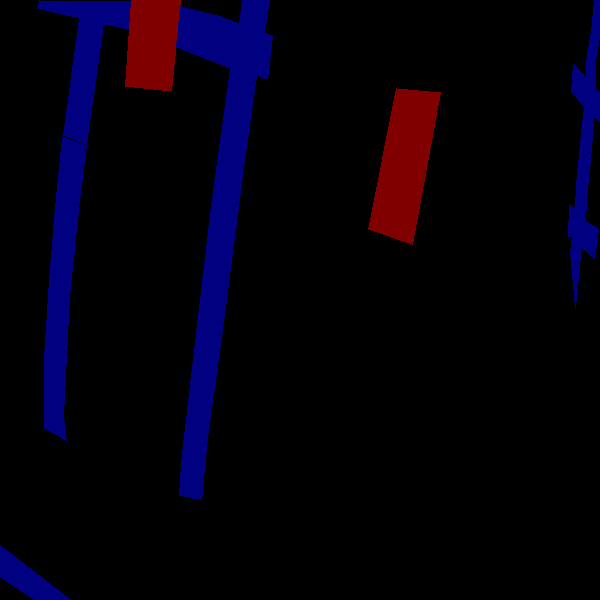}
        &
        \includegraphics[width=30mm,height=30mm]{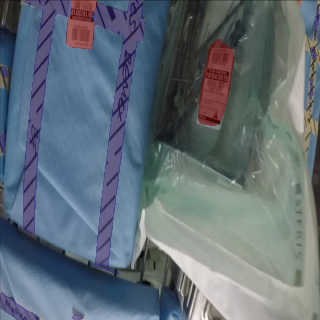}
        &
        \includegraphics[width=30mm,height=30mm]{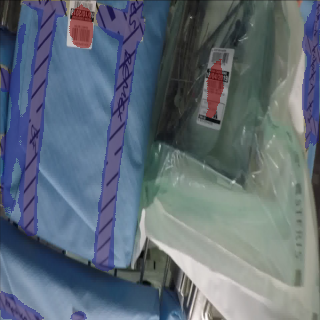}
        &
        \includegraphics[width=30mm,height=30mm]{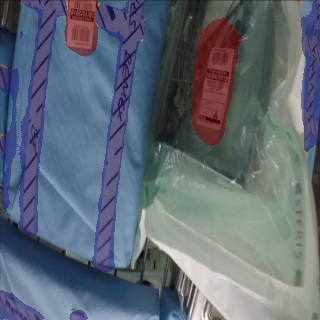}
        \\
        
        \includegraphics[width=30mm,height=30mm]{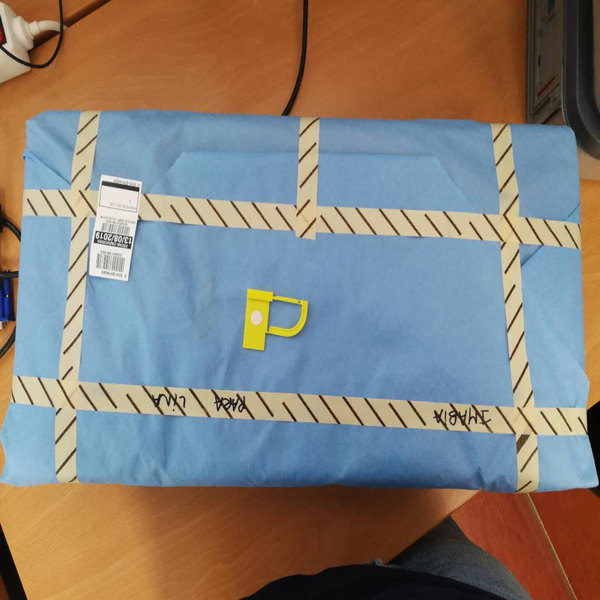}
        &
        \includegraphics[width=30mm,height=30mm]{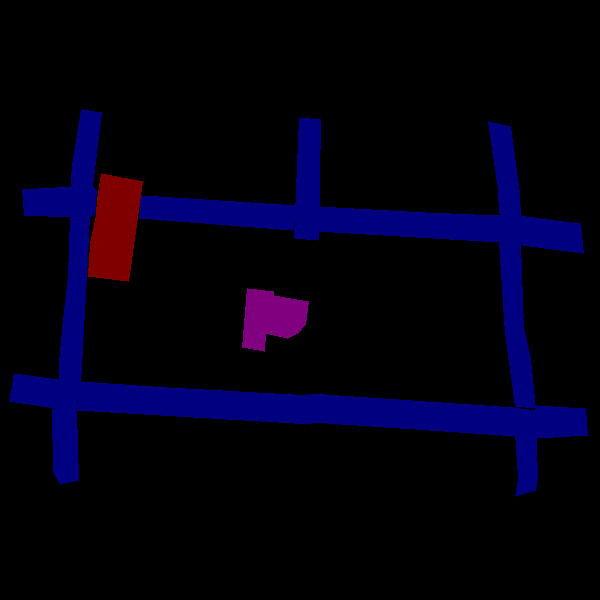}
        &
        \includegraphics[width=30mm,height=30mm]{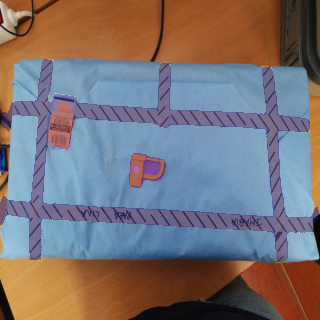}
        &
        \includegraphics[width=30mm,height=30mm]{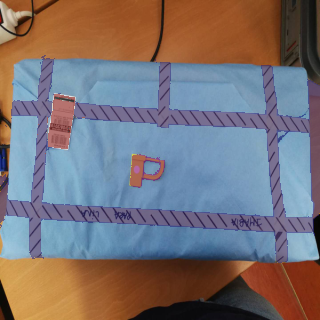}
        &
        \includegraphics[width=30mm,height=30mm]{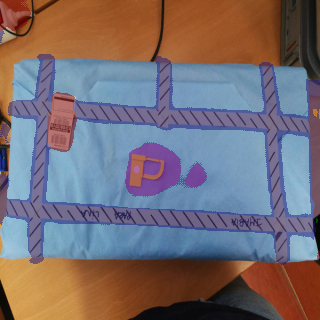}
        \\
        
        \includegraphics[width=30mm,height=30mm]{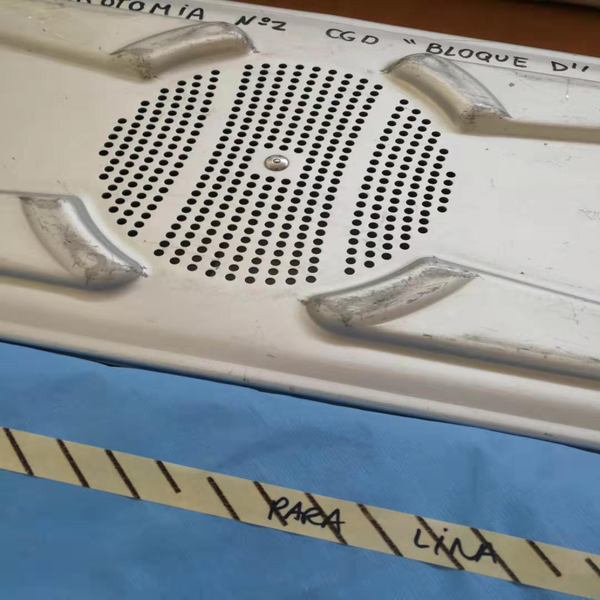}
        &
        \includegraphics[width=30mm,height=30mm]{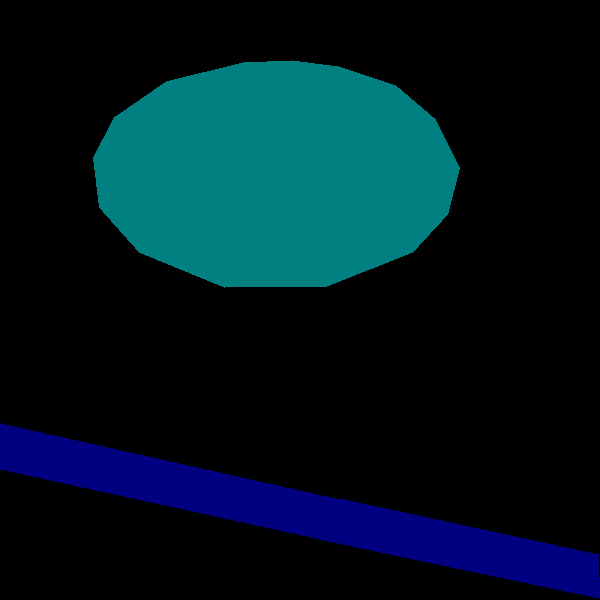}
        &
        \includegraphics[width=30mm,height=30mm]{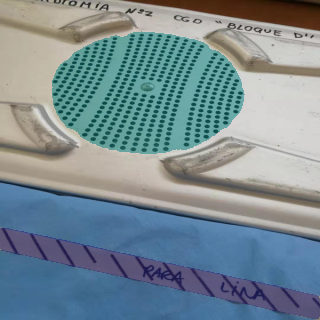}
        &
        \includegraphics[width=30mm,height=30mm]{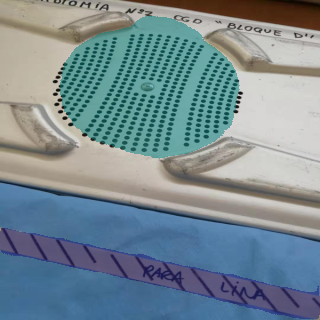}
        &
        \includegraphics[width=30mm,height=30mm]{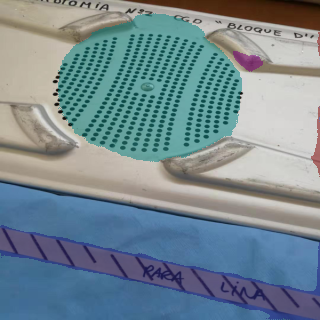}
        \\
        
        \includegraphics[width=30mm,height=30mm]{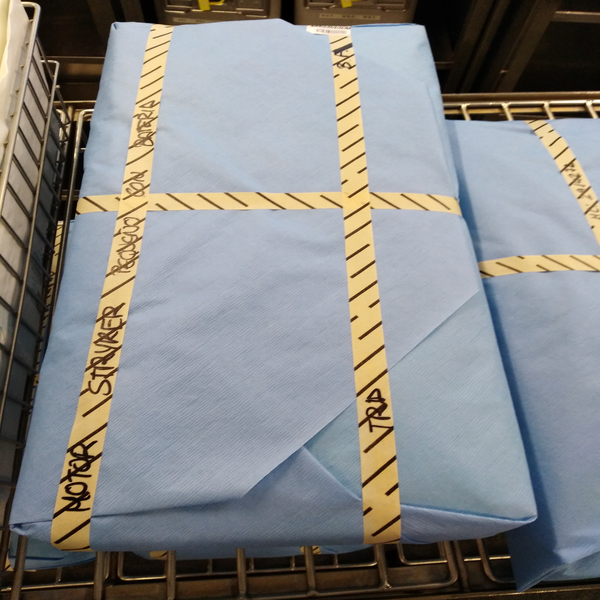}
        &
        \includegraphics[width=30mm,height=30mm]{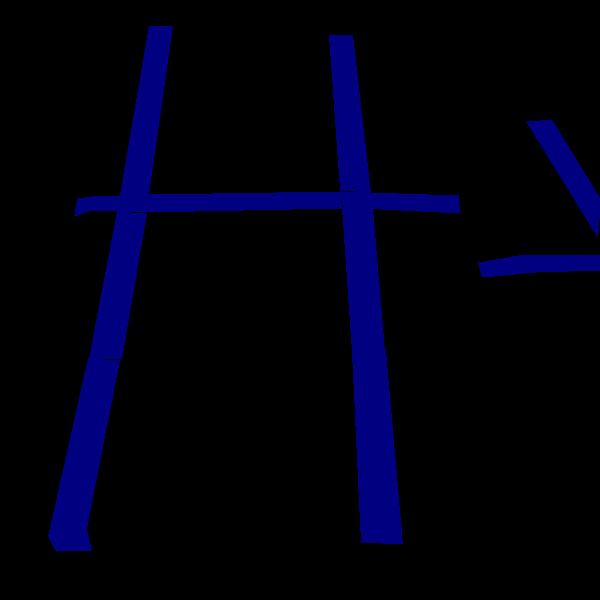}
        &
        \includegraphics[width=30mm,height=30mm]{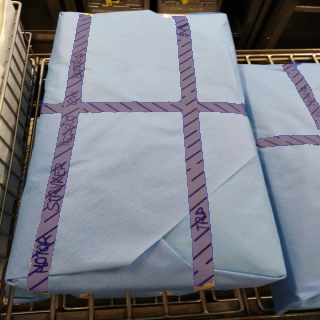}
        &
        \includegraphics[width=30mm,height=30mm]{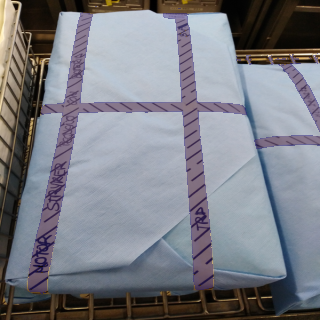}
        &
        \includegraphics[width=30mm,height=30mm]{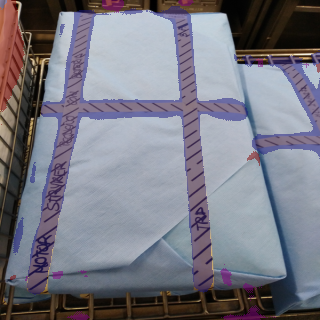}
        \\
        % Image & Ground Truth & Full Supervision & Clustering & Segmentation
        \footnotesize original image & 
        \footnotesize full mask & 
        \footnotesize E-FULL & 
        \footnotesize E-SCR20-SUP80-N (seg) & 
        \footnotesize E-SCR20-SUP80-N (clu)
    \end{tabular}
    %\vspace{-2mm}
    %\caption{Examples of segmentation results on the Quality Control dataset. The input images are shown in the first column; the second column shows the fully supervised ground truth; the third column shows the results of the fully supervised approach; the fourth column represents examples of the clustering results of E-SCR20-SUP80-N; its segmentation results obtained after dense CRF are shown in the fifth column. }
    \caption{Examples of segmentation results for the quality control task: (1st column) original images, (2nd column) full mask, (3rd column) results of the fully supervised approach, (4th column) segmentation output from E-SCR20-SUP80-N after dense CRF, (5th column) segmentation output from clustering for the same configuration.}
    \label{fig:box_seg_clu_results}
\end{figure*}

Figure~\ref{fig:seg_results} shows examples of segmentation results for the visual inspection task. As can be observed, the segmentations resulting from our approach are very similar to those from the upper baseline (E-FULL). Moreover, as expected, results from clustering are basically correct though tend to label incorrectly pixels (false positives) from around correct labellings (true positives). Similar conclusions can be drawn for the quality control task, whose results are shown in Fig.~\ref{fig:box_seg_clu_results}.

Summing up, the use of the Centroid Loss has made possible training a semantic segmentation network using a small number of labelled pixels. Though the performance of the approach is inferior to that of a fully supervised approach, the resulting gap for the two tasks considered has turned out to be rather short, given the challenges arising from the use of weak annotations.

\begin{sidewaystable} % [htbp] % htb
    \centering
    \caption{Segmentation results for the full loss function (G3). \textit{Seg} denotes that the segmentation output comes directly from the segmentation network, while \textit{Clu} denotes that the segmentation output is obtained from clustering. *CRF refers to the performance (mIOU) after dense CRF post-processing. Best performance is highlighted in bold.}
    \label{tab:annotation_compare}
    %{ p{2.cm}|p{4.5cm}||p{1.5cm}||p{1.cm}p{1.cm}p{1.cm}||p{1.5cm}|p{2.cm}||p{1.5cm}|p{2.cm}   }
    %\begin{tabular}{ p{1.5cm}|p{4.5cm}||p{1.5cm}|p{2.cm}|p{2.cm}|p{2.cm}||p{1.8cm}|p{1.8cm}p{1.8cm}||p{2.cm} }
    \small
    \begin{tabular}{c|c||c||c|c|c|c||c|c|c|c||c}
        \toprule
        Task & Experiments & wmIOU & mIOU (seg)      & mRec (seg)      & mPrec (seg)     & F$_1$ (seg) & mIOU (clu)      & mRec (clu)      & mPrec (clu)  & F$_1$ (clu) & *CRF (seg)     \\
        \hline
        \multirow{9}{1.5cm}{Visual Inspection}
        & E-SCR2-N        & 0.2721    & 0.6995          & 0.6447          & 0.6452          & 0.6449 & 0.6828          & 0.7663          & 0.5803          & 0.6605 & 0.7068          \\  
        & E-SCR5-N        & 0.2902    & 0.7134          & 0.6539          & 0.6542          & 0.6540 & 0.7001          & 0.7447          & 0.6015          & 0.6655 & 0.7212          \\  
        & E-SCR10-N       & 0.3074    & 0.7047          & 0.6797          & 0.6332          & 0.6556 & 0.6817          & 0.7741          & 0.5772          & 0.6613 & 0.7241          \\ 
        & E-SCR20-N       & 0.3233    & 0.6904          & 0.6917          & 0.6081          & 0.6472 & 0.6894          & \textbf{0.7816} & 0.5507          & 0.6461 & 0.7172          \\ 
        \cline{2-12}
        & E-SCR20-SUP30-N & 0.6272    & 0.6919          & \textbf{0.7937} & 0.7081          & 0.7485 & 0.6987          & 0.7806          & 0.5946          & 0.6750 & 0.7489          \\ 
        & E-SCR20-SUP50-N & 0.6431    & \textbf{0.7542} & 0.7543          & \textbf{0.7567} & \textbf{0.7555} & \textbf{0.7491} & 0.7725          & \textbf{0.6830} & \textbf{0.7250} & \textbf{0.7859} \\ 
        & E-SCR20-SUP80-N & 0.6311    & 0.7294          & 0.7452          & 0.7397          & 0.7424 & 0.7268          & 0.7758          & 0.6200          & 0.6892 & 0.7693          \\ 
        \cline{2-12}
        & E-FULL          & 1.0000    & 0.8333          & 0.8537           & 0.9119         & 0.8818 & -               & -               & -               & -      & 0.8218          \\
        \hline
        \multirow{4}{1.5cm}{Quality Control}
        & E-SCR20-SUP30-N & 0.4710    & 0.7030          & 0.7937          & 0.7924          & 0.7930 & 0.5910          & 0.7600          & 0.6798          & 0.7177 & 0.7142          \\ 
        & E-SCR20-SUP50-N & 0.5133    & 0.7291          & 0.8298          & \textbf{0.8439} & \textbf{0.8368} & 0.6605          & 0.7777          & 0.7332          & 0.7548 & 0.7143          \\ 
        & E-SCR20-SUP80-N & 0.5888    & \textbf{0.7679} & \textbf{0.8303} & 0.8398          & 0.8350 & \textbf{0.6687} & \textbf{0.8630} & \textbf{0.7606} & \textbf{0.8086} & \textbf{0.7707} \\ 
        \cline{2-12}
        & E-FULL          & 1.0000    & 0.8604          & 0.8058          & 0.8432          & 0.8241 & -               & -               & -               & -      & 0.8459          \\
        \bottomrule
    \end{tabular}
    % \end{adjustwidth}          
\end{sidewaystable}

\section{Conclusions and Future Work}
\label{sc:conclusions}

This paper describes a weakly-supervised segmentation approach based on Attention U-Net. The loss function comprises three terms, namely a partial cross-entropy term, the so-called Centroid Loss and a regularization term based on the mean squared error. They all are jointly optimized using an end-to-end learning model. Two industry-related application cases and the corresponding datasets have been used as benchmark for our approach. As has been reported in the experimental results section, our approach can achieve competitive performance, with regard to full supervision, with a reduced labelling cost to generate the necessary semantic segmentation ground truth. Under weak annotations of varying quality, our approach has been able to achieve good segmentation performance, counteracting the negative impact of the imperfect labellings employed.

The performance gap between our weakly-supervised approach and the corresponding fully-supervised approach has shown to be rather reduced regarding the mIOU values. As for precision and recall, they are quite similar for the quality control task for both the weakly-supervised and the fully-supvervised versions. A non-negligible difference is however observed for the visual inspection task, what suggests looking for alternatives even less sensitive to the imperfections of the ground truth deriving from the weak annotations, aiming at closing the aforementioned gap. In this regard, future work will focus on other deep backbones for semantic segmentation, e.g. DeepLab.

\bibliographystyle{unsrt}
\bibliography{abbrv,paper}

\begin{IEEEbiography}[{\includegraphics[width=1in,height=1.25in,clip,keepaspectratio]{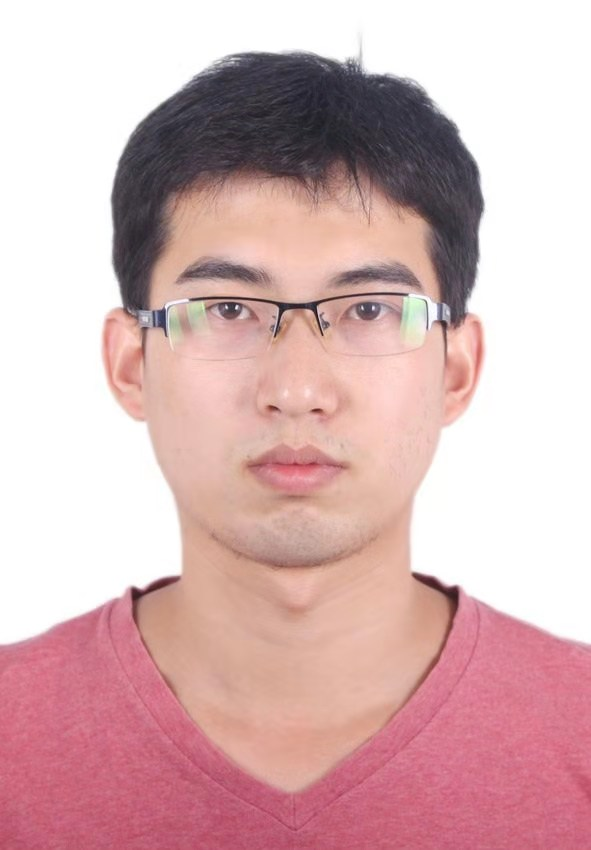}}]{Kai Yao} is currently pursuing a Ph.D. degree at the University of the Balearic  Islands (UIB). He received the B.Sc. and the M.Sc. degree from North China University of Technology. His current research interests include DCNN-based object detection and semantic segmentation, as well as the application of machine learning to visual inspection.
\end{IEEEbiography}

\begin{IEEEbiography}[{\includegraphics[width=1in,height=1.25in,clip,keepaspectratio]{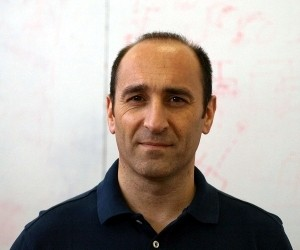}}]{Alberto Ortiz} is Associate Professor at the Department of Mathematics and Computer Science of the University of the Balearic Islands (UIB). He holds B.Sc. and Ph.D. degrees in Computer Science. He is author and co-author of more than 160 publications related with computer vision, machine learning and mobile robotics. His current research interests are machine learning (deep and shallow) and its applications, motion estimation, localization and mapping, visual guidance of mobile robots, including obstacle detection and avoidance, and control architectures for mobile robots.
\end{IEEEbiography}

\begin{IEEEbiography}[{\includegraphics[width=1in,height=1.25in,clip,keepaspectratio]{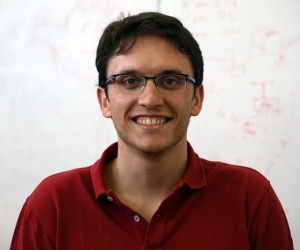}}]{Francisco Bonnin-Pascual} is a post-doctoral researcher and teaching assistant at the Department of Mathematics and Computer Science of the University of the Balearic Islands (UIB). He holds B.Sc., M.Sc. and Ph.D. degrees in Computer Science. His current research interests include motion estimation and control architectures for aerial robots, and computer vision and machine learning techniques applied to visual inspection.
\end{IEEEbiography}

\EOD

\end{document}